\documentclass[journal]{IEEEtran}

\usepackage{url}
\usepackage[printonlyused,nohyperlinks]{acronym} 
\usepackage{latexsym}
\usepackage[algo2e]{algorithm2e}
\usepackage{comment}
\usepackage{ifpdf}
 \ifpdf
 \else
 \fi
\usepackage{comment}
\usepackage{cite}
\ifCLASSINFOpdf
  \usepackage[pdftex]{graphicx}
  \graphicspath{{./pdf/}{./jpeg/}{./png/}}
  \DeclareGraphicsExtensions{.pdf,.jpeg,.png}
\else
  \usepackage[dvips]{graphicx}
  \graphicspath{{./eps/}}
  \DeclareGraphicsExtensions{.eps}
\fi

\usepackage[cmex10]{amsmath}
\usepackage{amsfonts}
\usepackage{amssymb}
\usepackage{amsthm}
\usepackage{subfig}
\usepackage{epstopdf}
\usepackage{pdfpages}
\usepackage{float}
\usepackage{color}

\usepackage{algorithmic}
\usepackage{array}
\usepackage{url}

\usepackage{stfloats}

\fnbelowfloat

\theoremstyle{plain}

\theoremstyle{definition}

\theoremstyle{remark}

\newcommand{\argmin}{\arg\!\min}

\newcommand{\abs}[1]{\left|#1\right|}

\begin{document}

\title{Cluster Naturalistic Driving Encounters Using Deep Unsupervised Learning}

\author{Sisi Li, Wenshuo Wang, Zhaobin Mo and Ding Zhao
\thanks{S. Li is with the Robotics Institute, University of Michigan, Ann Arbor, MI, 48109.}
\thanks{W. Wang is with the Department of Mechanical Engineering, University of Michigan, Ann Arbor, 48109.}
\thanks{Z. Mo is with the Automotive Engineering at the Tsinghua University, Beijing, China, 100084.}%
 \thanks{D. Zhao is with the Department of Mechanical Engineering, University of Michigan, Ann Arbor, 48109 (corresponding author: Ding Zhao zhaoding@umich.edu).}
\thanks{This work was funded by the Toyota Research Institute with grant No. N021936.}
}

\maketitle
\begin{abstract}
Learning knowledge from driving encounters could help self-driving cars make appropriate decisions when driving in complex settings with nearby vehicles engaged. This paper develops an unsupervised classifier to group naturalistic driving encounters into distinguishable clusters by combining an auto-encoder with $k$-means clustering (AE-$k$MC). The effectiveness of AE-$k$MC was validated using the data of 10,000 naturalistic driving encounters which were collected by the University of Michigan, Ann Arbor in the past five years. We compare our developed method with the $k$-means clustering methods and experimental results demonstrate that the AE-$k$MC method outperforms the original $k$-means clustering method. 
\end{abstract}

\begin{IEEEkeywords}
Driving encounter classification, unsupervised learning, auto-encoder
\end{IEEEkeywords}


\IEEEpeerreviewmaketitle

\section{Introduction}
Driving encounter in this paper is referred to as the scenario where two or multiple vehicles are spatially close to and interact with each other when driving. Autonomous vehicle has been becoming a hot topic in both industry and/or academia over past years and making human-like decisions with other encountering human drivers in complex traffic scenarios brings big challenges for autonomous vehicles. Currently, one of the most popular approaches to deal with the complex driving encounters is to manually and empirically classify it into several simple driving scenarios according to their specific application. For example, lane change behavior for autonomous driving applications were empirically decomposed into different categories\cite{nilsson2016if,nilsson2017lane} based on human driver's active-passive decision behavior\cite{do2017human}. Then machine learning techniques, such as Markov decision process (MDP) and partially observable MDP (POMDP)\cite{hubmann2017decision,kuderer2015learning}, were introduced to learn specific decision-making models. A well-trained model usually requires sufficient amounts of naturalistic driving data\cite{wang2017much}. The authors in \cite{gametheory} combined reinforcement learning with game theory to develop a decision-making controller for autonomous vehicles using the vehicle interactions data generated from a traffic simulator. However, the trained controller in \cite{gametheory} may become invalid when deployed in real world traffic environment. In addition, directly feeding the training data into model without any classification could not make full use of underlying data resource and thereby could miss some rare but important driving scenarios in real traffic settings. One of the most significant challenges in solving this issue is to obtain massive high-quality labeled and categorized traffic data from naturalistic traffic settings, which is essential for the robustness and accuracy of learning decision-making models. 

Modern sensing technologies such as cameras, Lidar, and radar give great advantages to gather large-scale traffic data with multiple vehicles encountering; for instance, the University of Michigan Transportation Research Institute (UMTRI) equipped multiple buses with GPS tracker in Ann Arbor to gather the vehicle interactions data for autonomous vehicle research and more released database listed in\cite{wang18extracting, zhao2017trafficnet}. The gathered vehicles' interaction data could be explored to develop autonomous vehicle controller and generate testing motions to evaluate self-driving algorithms \cite{zhao2017accelerated, zhao2017carfollow, huang2017accelerated}. However, how to label such a massive dataset efficiently and group driving encounters in a reasonable way still remain as challenges. Labeling a huge amount of data manually is a time-consuming task, which requires the data analysts with rich prior knowledge covering the fields of traffic, intelligent vehicles as well as human factors. Ackerman claimed that\footnote[1]{\url{https://spectrum.ieee.org/cars-that-think/transportation/self-driving/how-driveai-is-mastering-autonomous-driving-with-deep-learning}} for one-hour well-labeled training data, it approximately takes 800 human hours. Ohn-Bar\cite{ohn2017all} took excessive time to manually annotate the objects in recorded driving videos to investigate the importance rank of interesting objects when driving. In order to meet the data-hungry learning-based methodologies, the industry and academia both need a tool that could automate the labeling process, thereby effectively eliminating labeling costs\cite{appenzeller2017scientists}. Except for classifying driving encounters, research has been conducted in recent years to automate the data labeling process, for example, ranging from labeling the ambiguous tweets automatically using supervised learning methods\cite{super_tweets} to classifying driving styles using unsupervised learning methods (e.g., $k$-means)\cite{martinez2017driving} and semi-supervised learning methods (e.g., semi-supervised support vector machine)\cite{wang2017driving}. Although these aforementioned unsupervised and semi-supervised approaches could reduce labeling efforts, but they are not suitable to deal with huge amounts of high-dimensional time-series data. Toward this end, the deep learning approach is becoming popular to gain insight into such data. For example, the auto-encoder and its extensions have been used to analyze driving styles\cite{dong2017autoencoder} and driver behavior \cite{liu2017visualization}, which empirically demonstrates its effectiveness. For autonomous vehicles, learning models from these manually predefined categories could suffer the following limitations: 

\begin{enumerate}
\item Manually labeling huge amounts of high-dimensional data requires excessive cost of time and resources and could generate large biases due to the diverse prior knowledge of data analysts.
\item The most suitable classified categories of driving encounters for human understanding may not generate the most suitable results to learn a decision-maker for self-driving cars.
\end{enumerate}

This paper presents an unsupervised framework, i.e., combining an autoencoder with $k$-means clustering (AE-$k$MC), to automatically cluster driving encounters with less subjective interference. The auto-encoder is employed as a component of extracting hidden features in a driving encounter classifier. The data were collected by the M-City with more than 2,800 cars, including commercial trucks and transit vehicles, throughout five years. The original contribution of this work is that an autoencoder-based framework was developed and implemented to automatically label the driving encounters according to the identified features, which has not been previously proposed elsewhere to the best of our knowledge. Finally, we make a comprehensive analysis for experiment results.

The rest of the paper is organized as follows. The developed AE-$k$MC approach is detailed in Section~\ref{sec: auto-encoder}. The vehicle encounter data collection and model training procedure are shown in Section~\ref{sec: model}. Section~\ref{sec: result} shows the experiment results. Section~\ref{sec: conclusion} presents discussion and conclusion.

\section{Unsupervised Learning Methods}
\label{sec: auto-encoder}
In this section, we will present two different unsupervised schemes to automatically cluster vehicle encounters using $k$-means clustering and AE-$k$MC. In what follows, we will introduce the theoretical basis of the traditional auto-encoder, the architecture of auto-encoder neural networks, and the $k$-means clustering method.


\subsection{Auto-Encoder Framework}
\label{sec: AE}

\begin{figure}[t]
\centering
\includegraphics[trim = 1cm 0cm 0cm 1cm,width=0.35\textwidth]{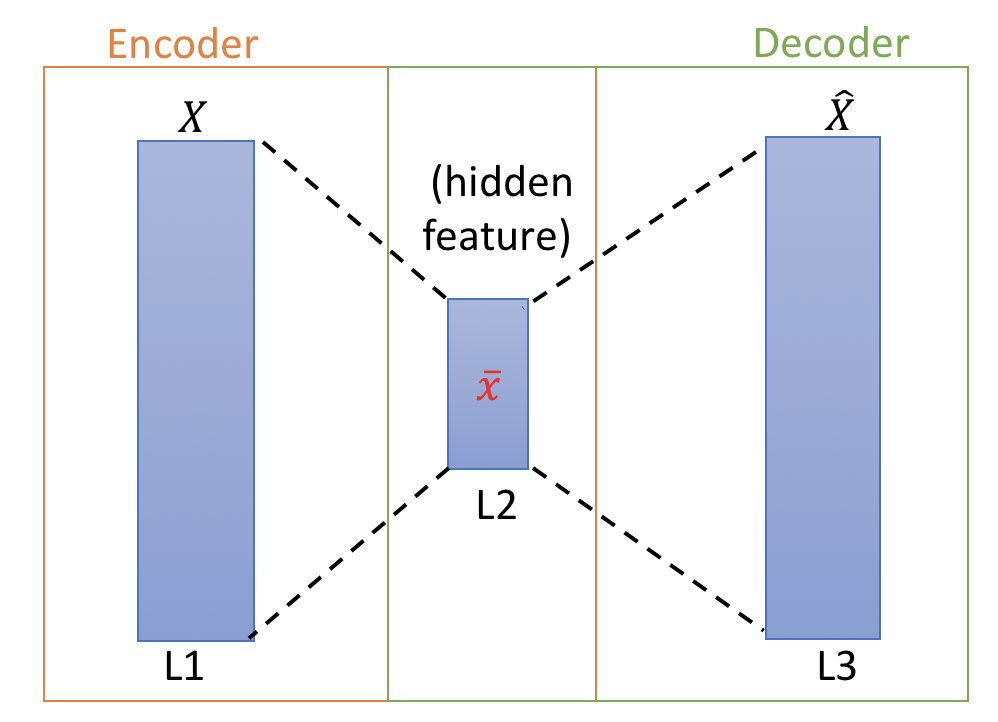}
\caption{Structure of a typical auto-encoder.}
\label{fig: common_auto_encoder_diag}
\end{figure}

The auto-encoder implemented in this work is a typical auto-encoder that contains a encoder and a decoder, as shown in Fig.~\ref{fig: common_auto_encoder_diag}.

\subsubsection{Encoder}
\label{sec: encoder}

\begin{figure}[t]
\centering
\includegraphics[trim = 1cm 0cm 0cm 1cm, width=0.4\textwidth]{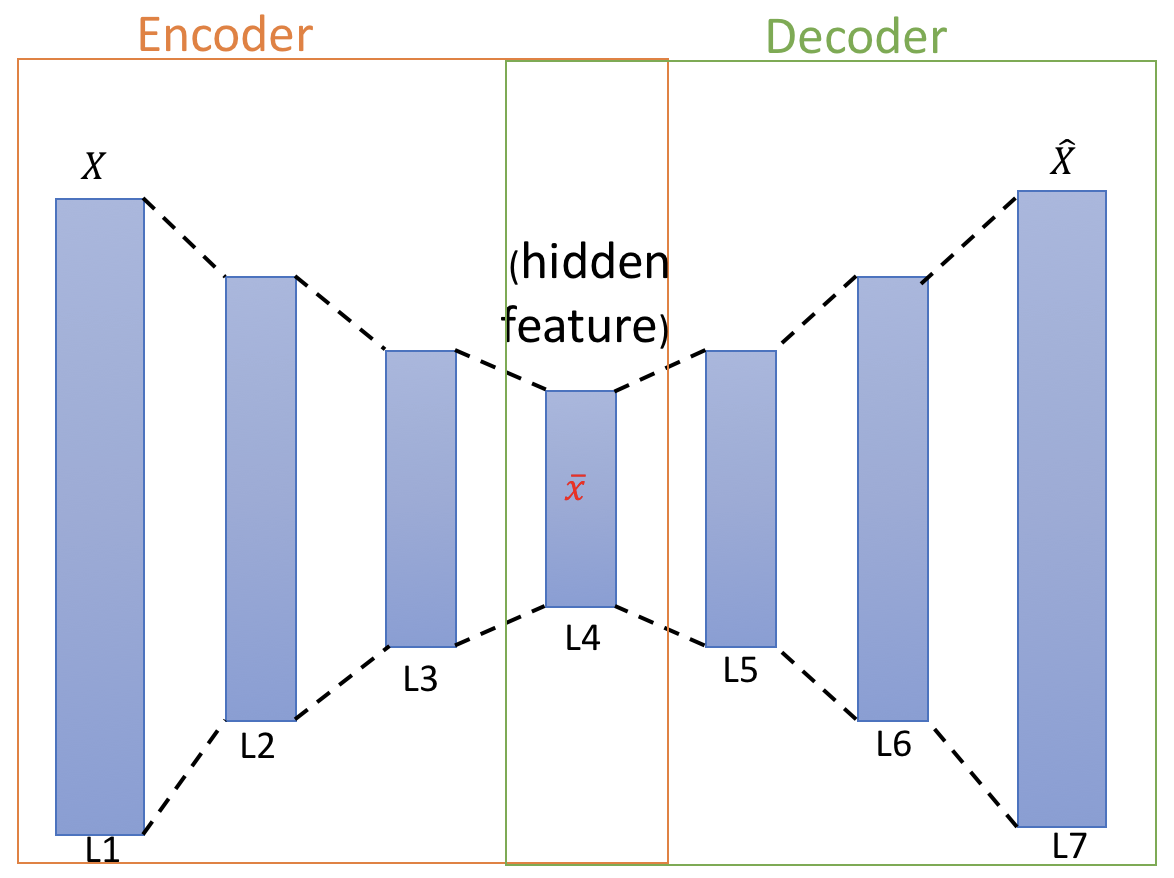}
\caption{Structure of the auto-encoder used in this paper.}
\label{fig: auto_encoder_diag}
\end{figure}

The encoder, labeled by an orange box in Fig.~\ref{fig: common_auto_encoder_diag}, maps an input data, $\mathbf{X} \in \mathbb{R}^{n}$, into a hidden representation $\mathbf{x} \in \mathbb{R}^{m}$ \cite{stacked}: 

\begin{subequations}
\begin{gather}
f_{\theta}(\mathbf{X}) = \mathbf{W}\mathbf{X} + \mathbf{b}, 
\end{gather}
\end{subequations}
where $ \mathbf{W} \in \mathbb{R}^{m \times  n}$ is a weight matrix and $ \mathbf{b} \in \mathbb{R}^{m}$ is a bias term. Unlike the normal encoder that only has one input layer and one output layer, the encoder we used is a four-layer neural network that contains one input layer (L1), two hidden layers (L2 and L3) and one output layer (L4), as shown in Fig.~\ref{fig: auto_encoder_diag}. Two hidden layers are added because each of those two extra layers represents an approximation of any function according to the universal approximator theorem \cite{goodfellow2016deep}. The modified encoder benefits for the new depth since the training cost, both computational cost and training data size, can be reduced \cite{goodfellow2016deep}. The representation of the data at each layer is calculated as:

\begin{equation}
    \mathbf{X_{i+1}} = \mathbf{W_{i,i+1}}\mathbf{X_i} + \mathbf{b_{i,i+1}}, 
\end{equation}
where $\mathbf{X_i}$ is the input data representation at layer $L_i$, and $\mathbf{W_{i,i+1}}$ and $\mathbf{b_{i,i+1}}$ are the weight vector and the bias vector, respectively, capable of mapping $\mathbf{X_i}$ to $\mathbf{X_{i+1}}$. Note that the data representation at the layer 4, $\mathbf{\bar{x}}$, is the code or the hidden feature of the input data. 

\subsubsection{Decoder}

The coded data, or the hidden representation $\bar{\mathbf{x}} \in \mathbb{R}^{n} $ from the encoder will be mapped back to $\hat{X}$ through layers L5 and L7, which is the reconstructed representation of the original input data $\mathbf{X}$. Note that the decoder employed in this work is a deep neural network that contains one input layer (L4), two hidden layers (L5 and L6) and one output layer.

\subsection{Optimization}

In order to reconstruct the input data from the hidden feature, $\mathbf{\bar{x}}$, the error between the reconstructed representation and the original input should be minimized. Hence the cost function can be defined as

\begin{equation}
    E(\mathbf{\Omega}) = (\mathbf{X} - \hat{\mathbf{X}} )^T(\mathbf{X} - \hat{\mathbf{X}} ),
\end{equation}
where $\mathbf{\Omega}$ is the parameter set that contains the weight vector and the bias vector of each layer, $\hat{\mathbf{X}} = g_{\Omega}(\mathbf{\bar{x}})$, $g_{\Omega}$ is the mapping that map the hidden representation $\mathbf{x}$ to $\hat{\mathbf{X}}$. Then, the optimization problem is defined as

\begin{equation}
    \Omega^* = \arg \min_\Omega E(\Omega).
\end{equation}

\subsection{$k$-Means Clustering}
The $k$-means clustering is one of popular unsupervised machine learning techniques for classification. Applying the trained auto-encoder on the raw vehicle encounter data, the hidden feature can be extracted and then fed into the $k$-means clustering. Given $n$ observations ($x_1$, $x_2$, $\dots$, $x_n$) and define $k$ classes, the $k$-means clustering method clusters the observations into $k$ groups by solving the optimization problem

\begin{equation} 
\mathbf{v}^* = \argmin_\mathbf{v} \sum_{i = 1} ^{k}\sum_{j = 1}^{n}\abs{x_{i} - v_{j}}^{2},
\end{equation}
where $\abs{x_{i} - v_{j}}$ is the Eucledian distance between the centroid and the observation \cite{kmeans}. The objective function tries to pick centroids that minimize the distances to all points belonging to its respective cluster so that the centroids are more representative of the surrounding cluster of data points.

\section{Data Collection and Experiment}
\label{sec: model}
\subsection{Data Collection}

\begin{figure}[ht]
\centering
\includegraphics[trim=1cm 1cm 3cm 1cm, clip=true,width=0.95\linewidth]{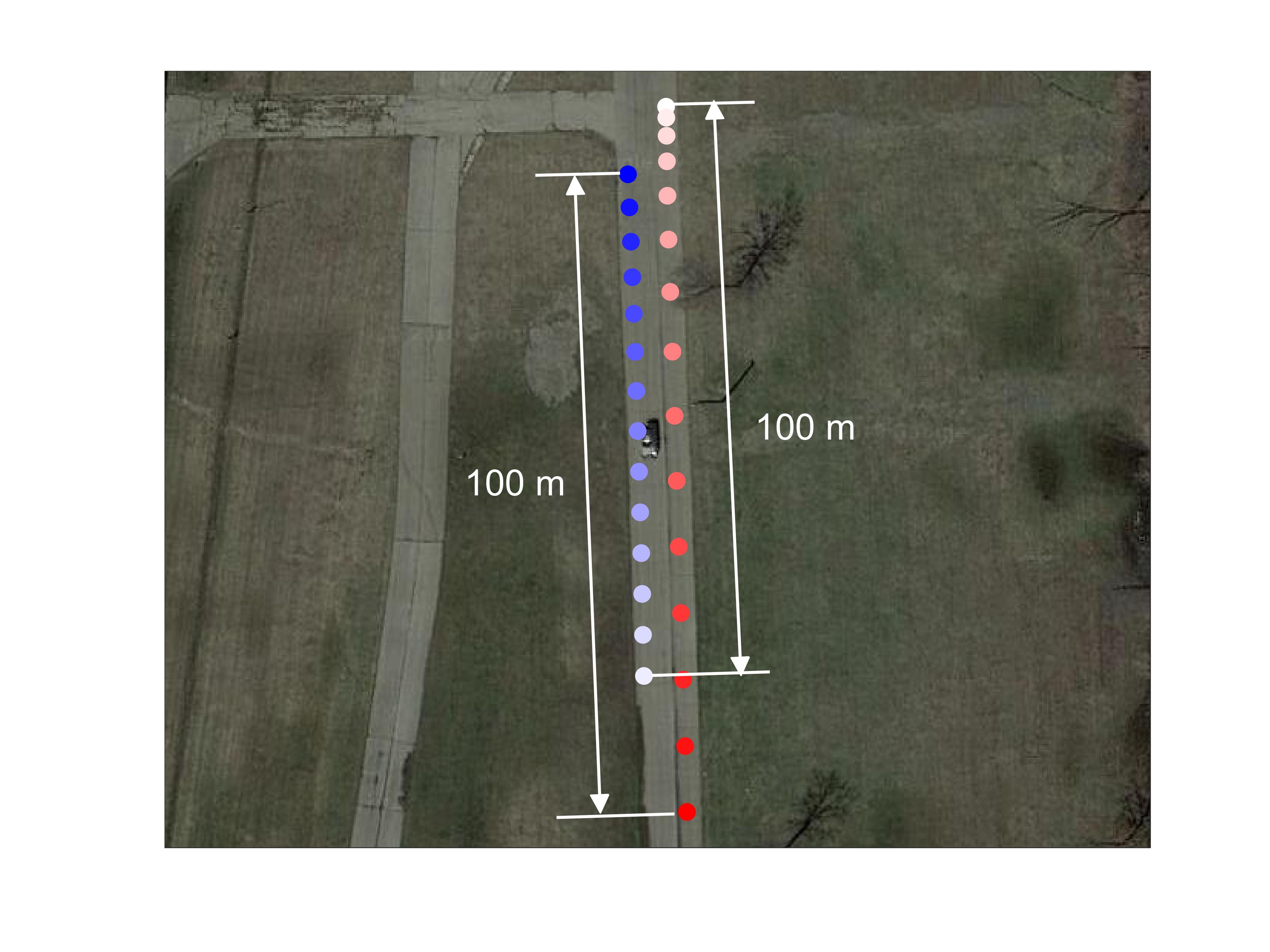}
     \caption{Example of driving encounters. Dark dot is start point and light dot is end point.}
     \label{fig:encounter}
 \end{figure}

We used the Safety Pilot Model Deployment (SPMD) database, which provides sufficient naturalistic data\cite{wang2017much}. This database was conducted by the University of Michigan Transportation Research Institute (UMTRI) and provides driving data logged in the last five years in Ann Arbor area \cite{huang2017empirical}. It includes approximately 3,500 equipped vehicles and 6 million trips in total. Latitude and altitude information for clustering was collected by the on-board GPS. The collection process starts at the ignition for each equipped vehicle. The data was collected with a sampling frequency of 10 Hz. 

We used the dataset of 100,000 trips, collected from 1900 vehicles with 12-day runs. The trajectory information we extracted includes latitude, longitude, speed and heading angle of the vehicles. The selection range was constricted to an urban area with the range of latitude and longitude to be (-83.82,-83.64) and (42.22,42.34), respectively. The vehicle encounter was defined as the scenario where the vehicle distance was small than 100 m, as shown in Fig. \ref{fig:encounter}. The dots indicate the position of the vehicle at every sample time. After querying from the SPMD database, we got 49,998 vehicle encounters. 

\begin{figure}[ht]
\centering
\includegraphics[width=0.95\linewidth]{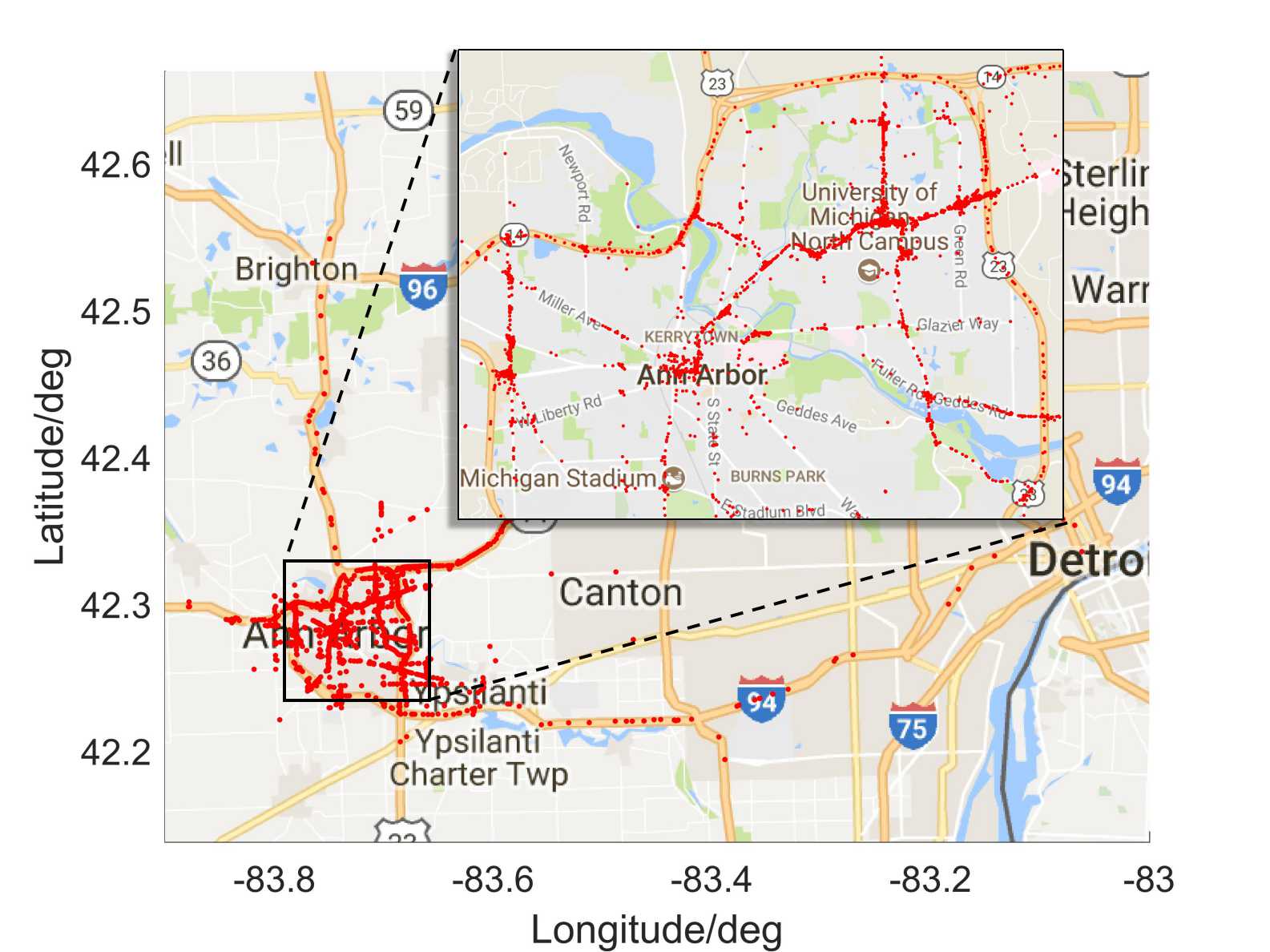}
     \caption{Distribution of driving encounters.}
     \label{fig:distri_encounter}
\end{figure}

The distribution of these encounters is shown in Fig. \ref{fig:distri_encounter}. The central points of the minimal rectangular that can encompass the trajectory is used to include the massive trajectories in one image. Note that to reduce the computational load, 10,000 encounters were randomly selected to test the unsupervised clustering methods.

\subsection{Model Training Procedure}
\begin{figure}[ht]
\centering
\includegraphics[trim=0.2cm 0.1cm 1.6cm 2cm, clip=true,width=0.52\textwidth]{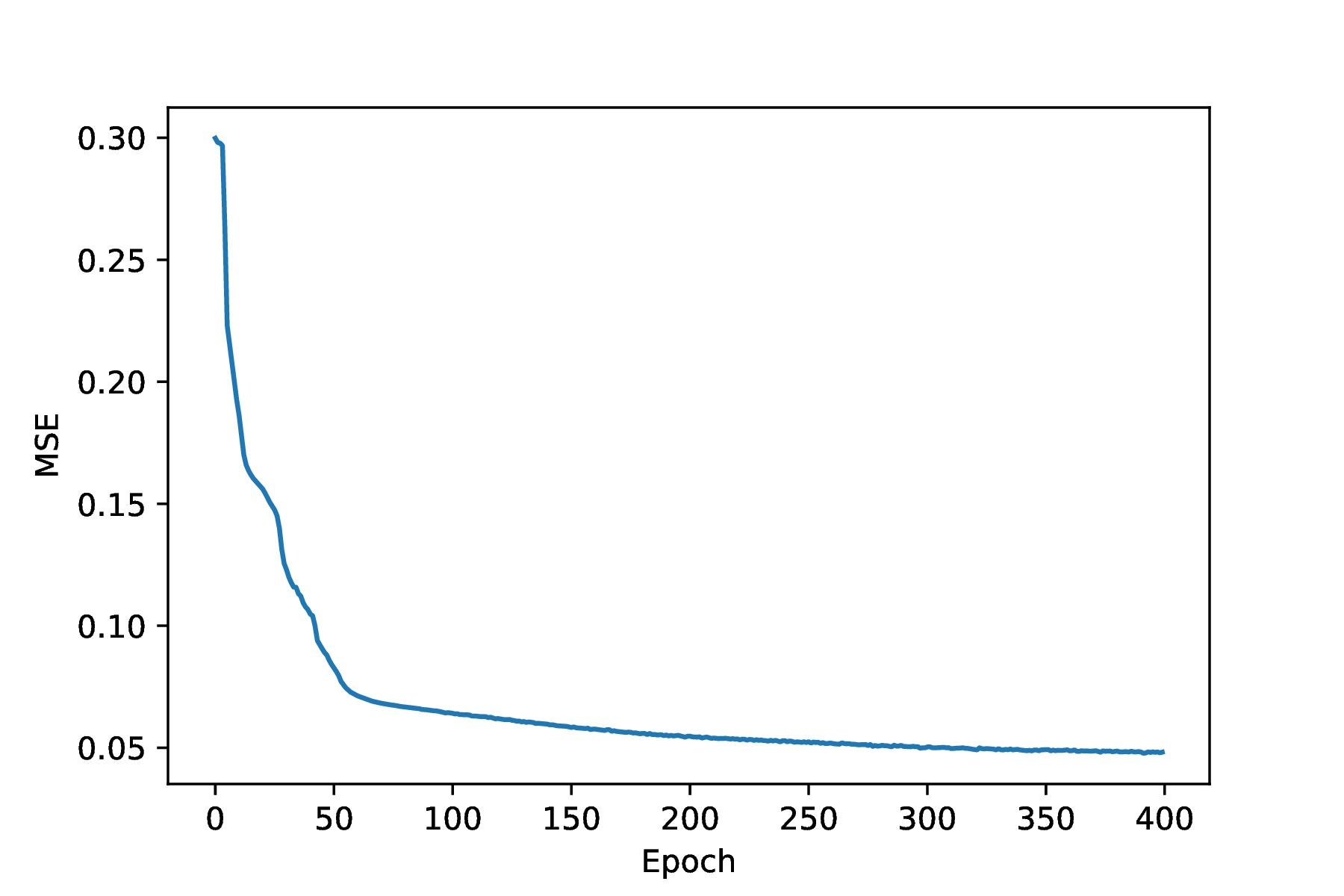}
\caption{Cost during the training of the model.}
\label{fig: cost}
\end{figure}

\begin{figure} [t] 
    \centering
  \subfloat[Two vehicles merge]{
       \includegraphics[trim=0.5cm 0.5cm 2.5cm 1cm, clip=true,width=0.45\linewidth]{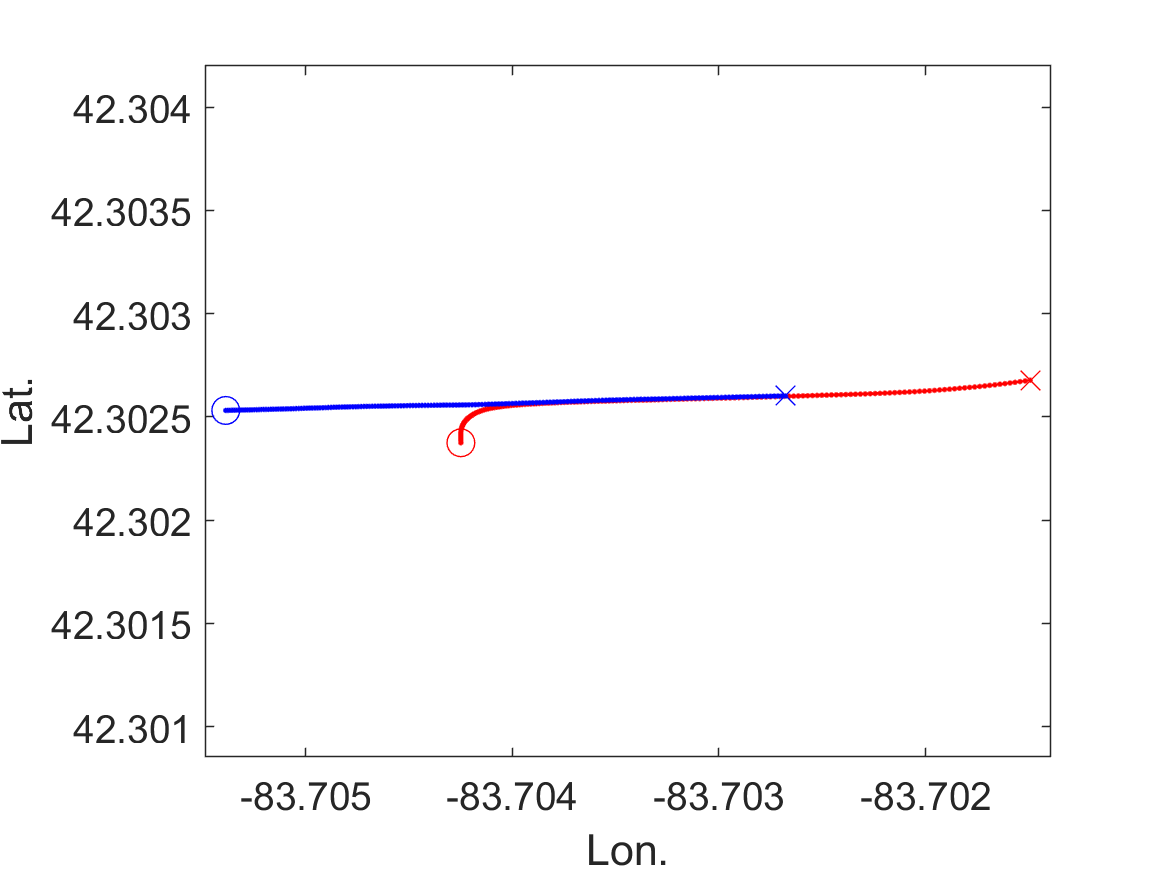}}\
  \subfloat[One vehicle follows another vehicle]{
        \includegraphics[trim=0.5cm 0.5cm 3cm 1cm, clip=true,width=0.45\linewidth]{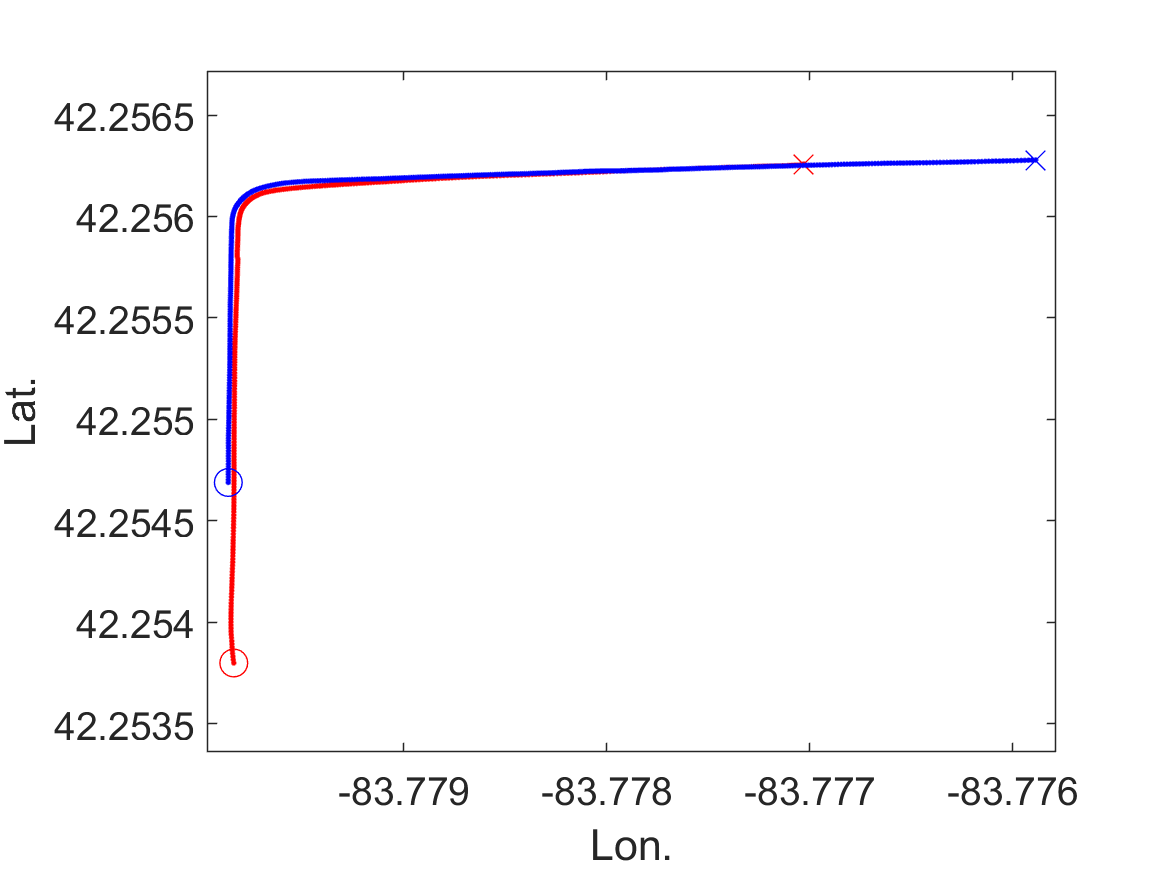}}\\
    \centering
 \subfloat[One vehicle bypasses another vehicle]{
       \includegraphics[trim=0.5cm 0.5cm 3cm 1cm, clip=true,width=0.45\linewidth]{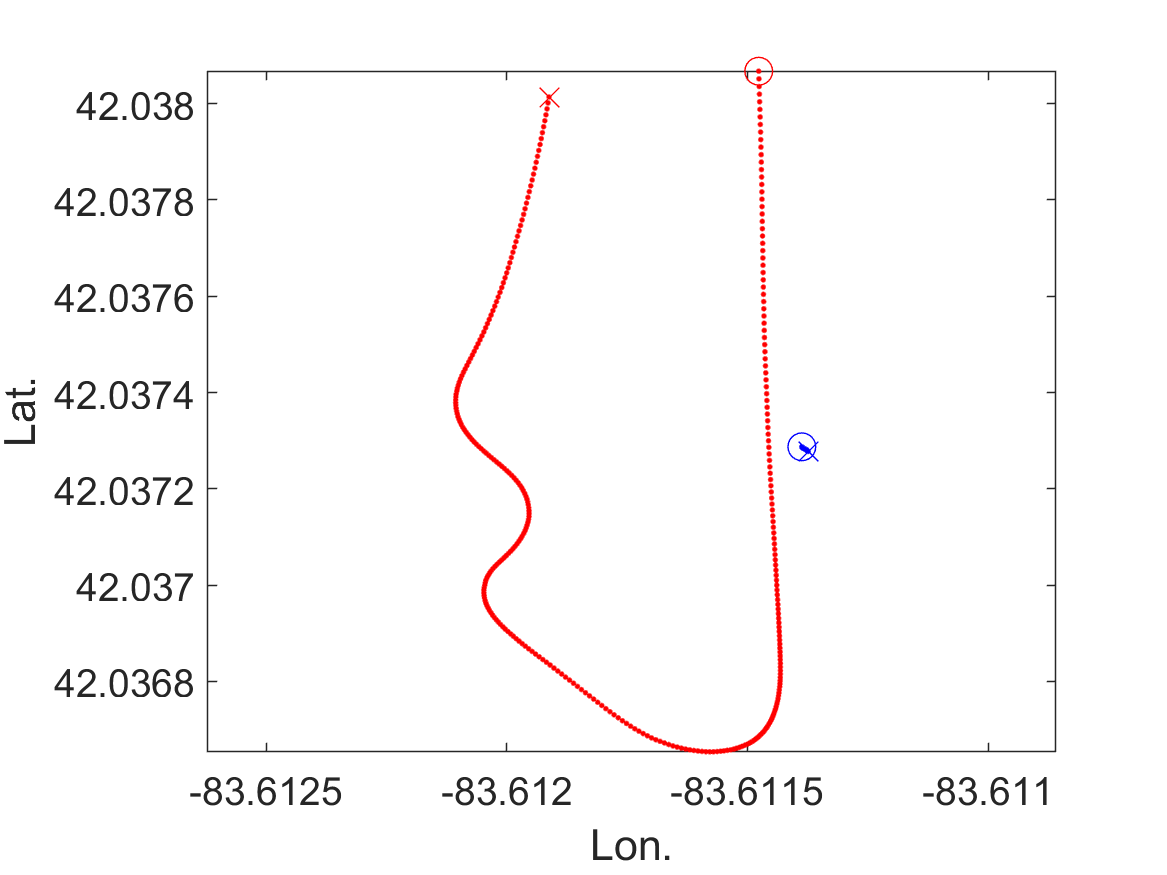}}\
  \subfloat[Two vehicles encounter in the same road from opposite directions]{
        \includegraphics[trim=0.5cm 0.5cm 2.8cm 1.2cm, clip=true,width=0.45\linewidth]{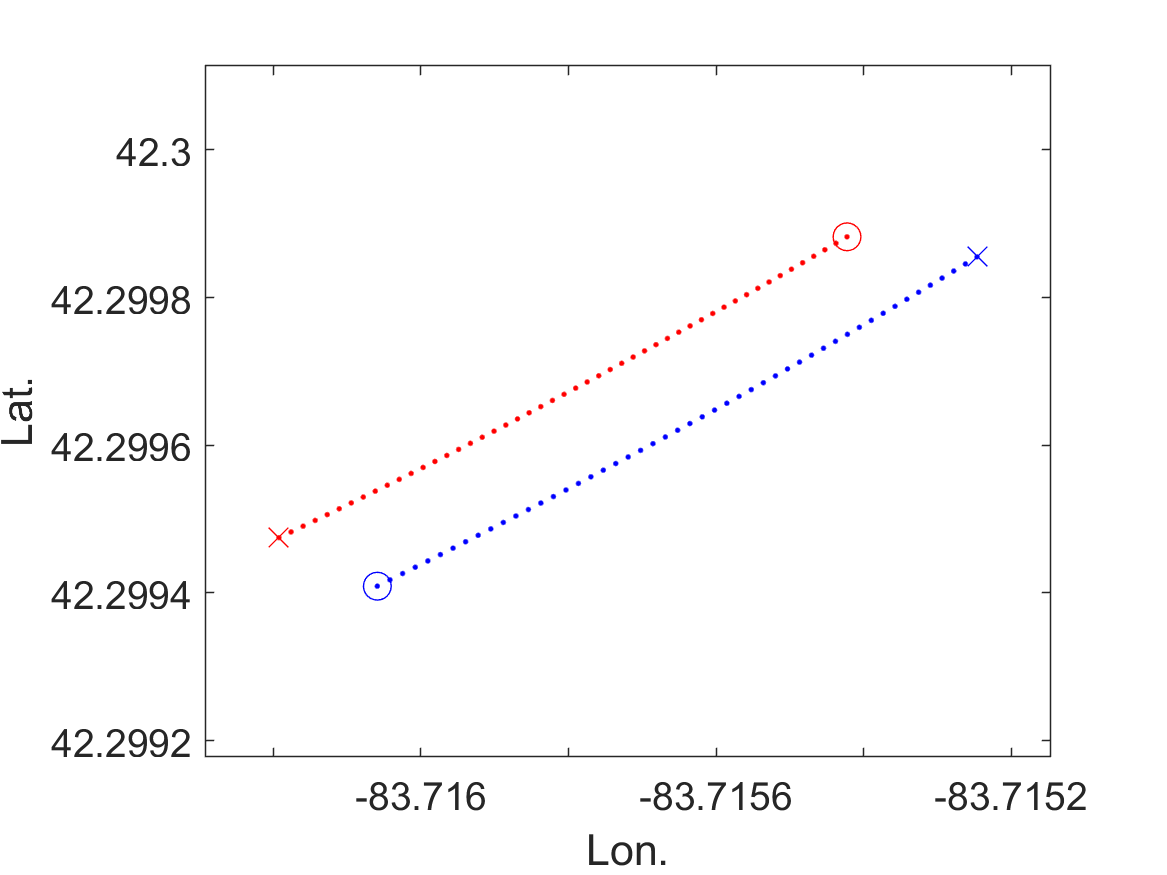}}
  \caption{Original vehicle interaction trajectories. Horizontal and vertical axises are the longitude and latitude of vehicles, respectively.}
  \label{fig: hidden_1} 
\end{figure}

The GPS data of an encounter was used as the input of the auto-encoder. Before doing this, the data was normalized to put all the training data in the same scale. The weights for each layers were initialized using random numbers between $1$ and $-1$ following uniform distribution. More specifically, we set $\mathbf{W_{1,2}} \in \mathbb{R}^{100  \times 200}$, $\mathbf{W_{2,3}} \in \mathbb{R}^{50 \times 100}$, $\mathbf{W_{3,4}} \in \mathbb{R}^{25 \times 50}$, $\mathbf{W_{4,5}} \in \mathbb{R}^{25  \times 50}$, $\mathbf{W_{5,6}} \in \mathbb{R}^{50 \times 100}$, $\mathbf{W_{6,7}} \in \mathbb{R}^{100 \times 200}$. The classic stochastic gradient decent method was employed to learn $\Omega$. The cost function value during the training process is shown in Fig.~\ref{fig: cost}. The training error converges to 0.041, which shows the trained Auto-encoder could reconstruct the input data from the hidden feature. Fig. \ref{fig: hidden_1} shows an example of the raw input of auto-encoders, i.e, raw vehicle GPS trajectory.

\section{Result Analysis and Discussion}
\label{sec: result}

\begin{figure} [ht]
    \centering
  \subfloat[\label{fig: int1}]{%
       \includegraphics[trim=5.5cm 2.9cm 3.5cm 2cm, clip=true, width=0.49\linewidth]{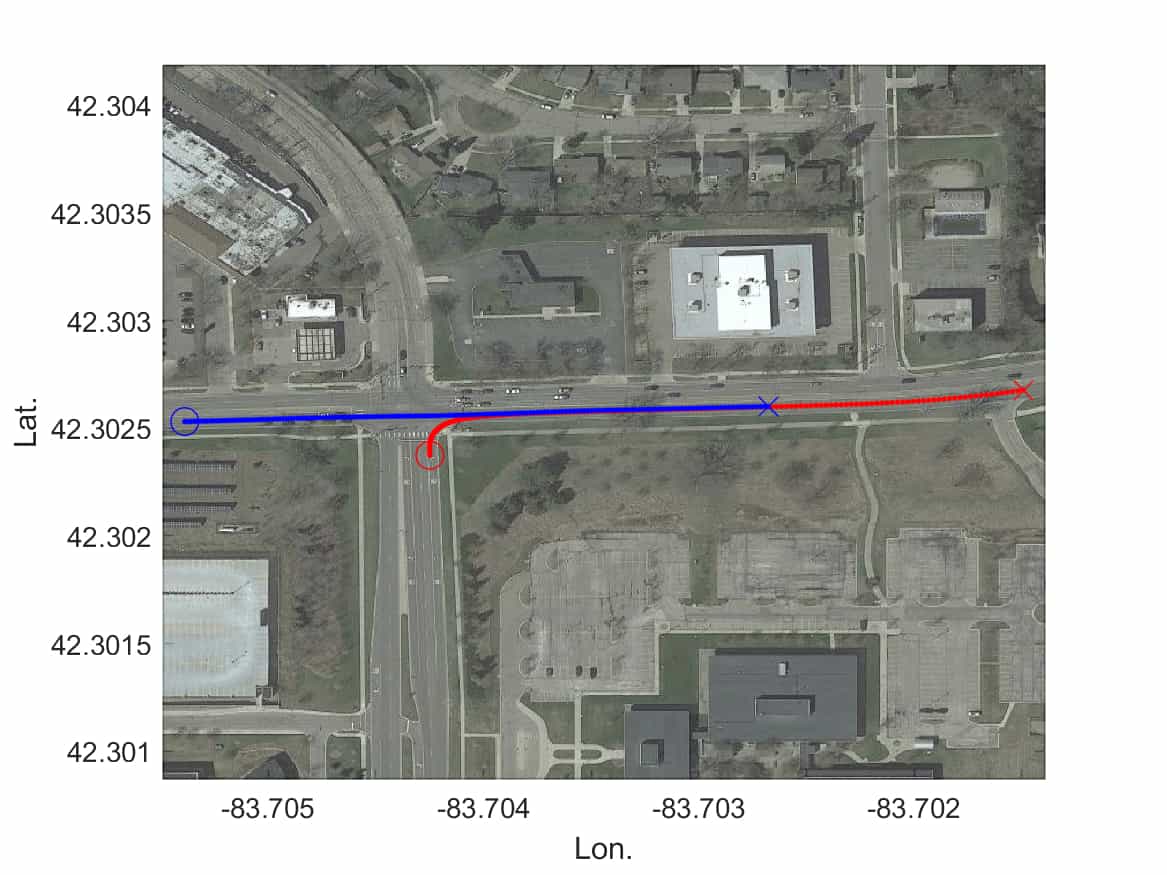}}\hfill
  \subfloat[\label{fig: int2}]{%
        \includegraphics[trim=5.5cm 2.9cm 3.5cm 2cm, clip=true,width=0.49\linewidth]{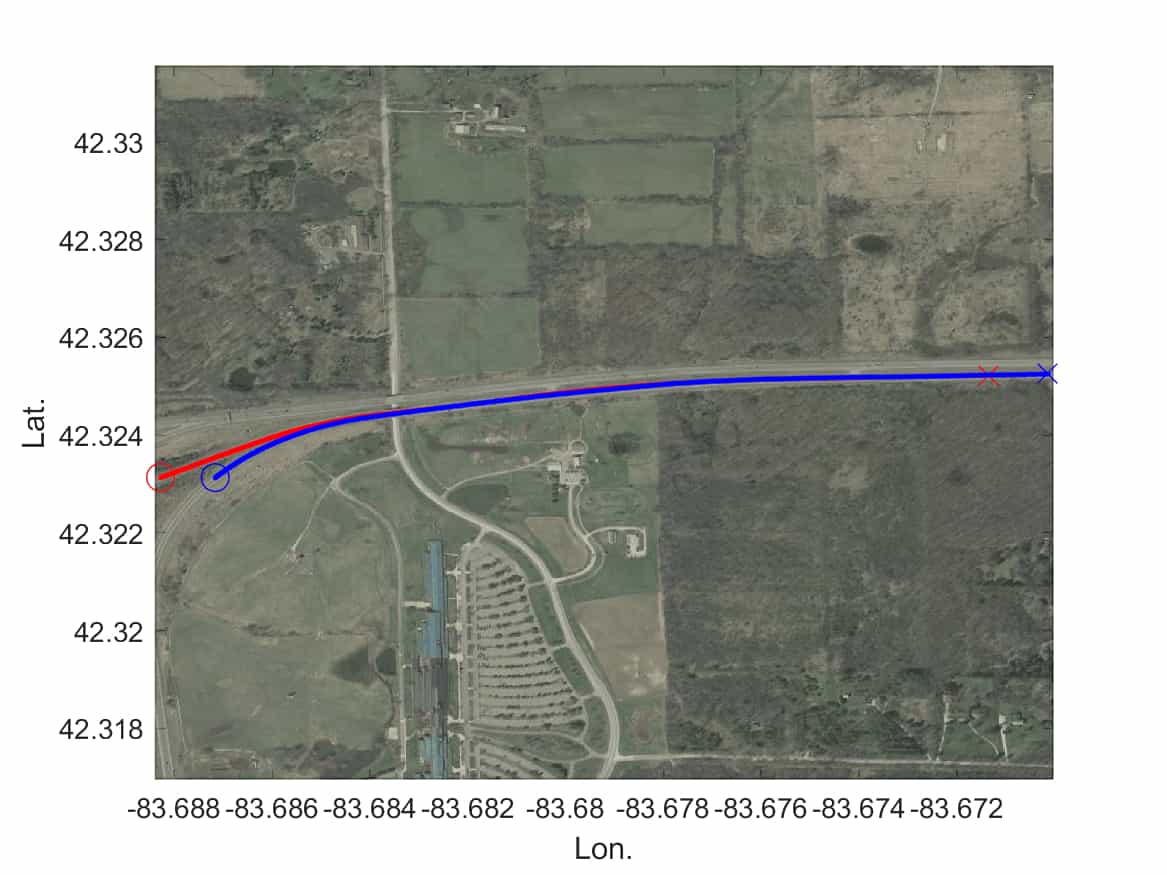}}\hfill
     \subfloat[\label{fig: int3}]{%
       \includegraphics[trim=5.5cm 2.9cm 4cm 2cm, clip=true,width=0.49\linewidth]{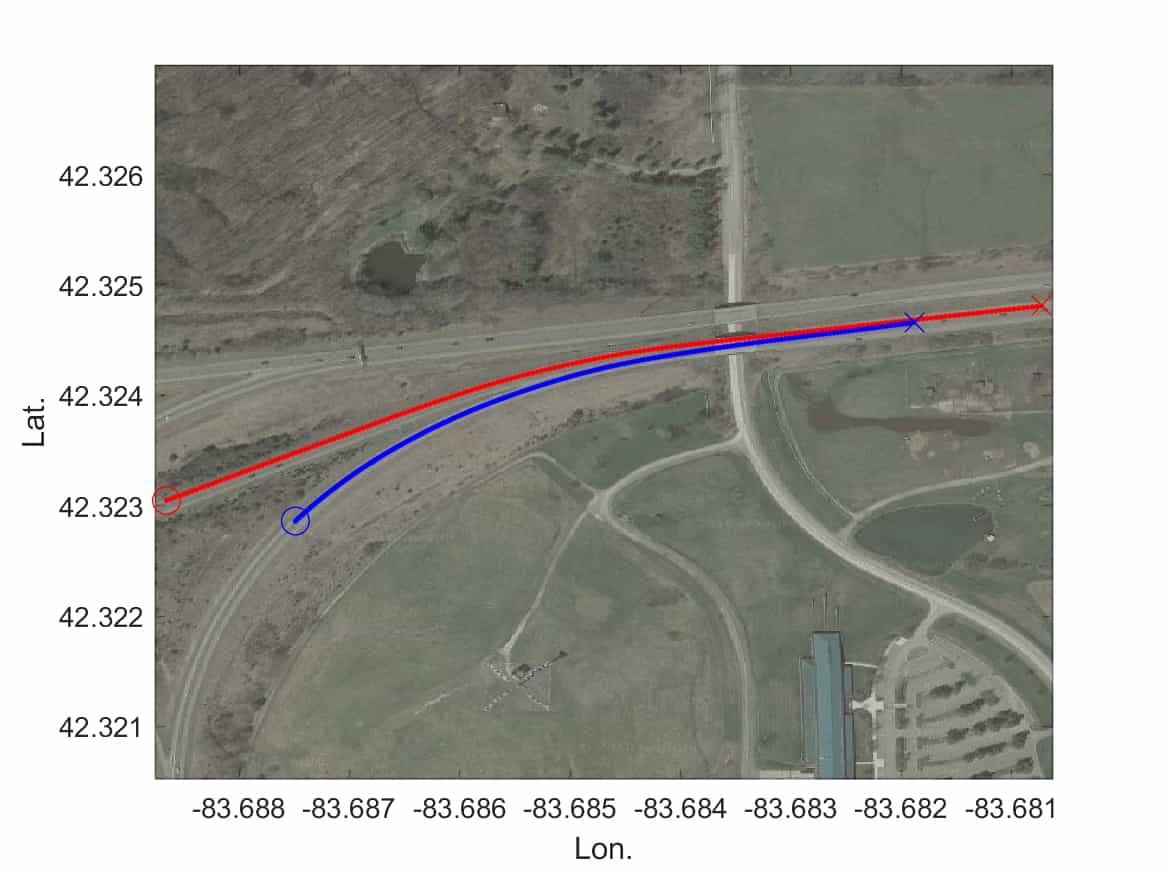}}\hfill
  \subfloat[\label{fig: int4}]{%
        \includegraphics[trim=5.5cm 2.9cm 4cm 2cm, clip=true,width=0.49\linewidth]{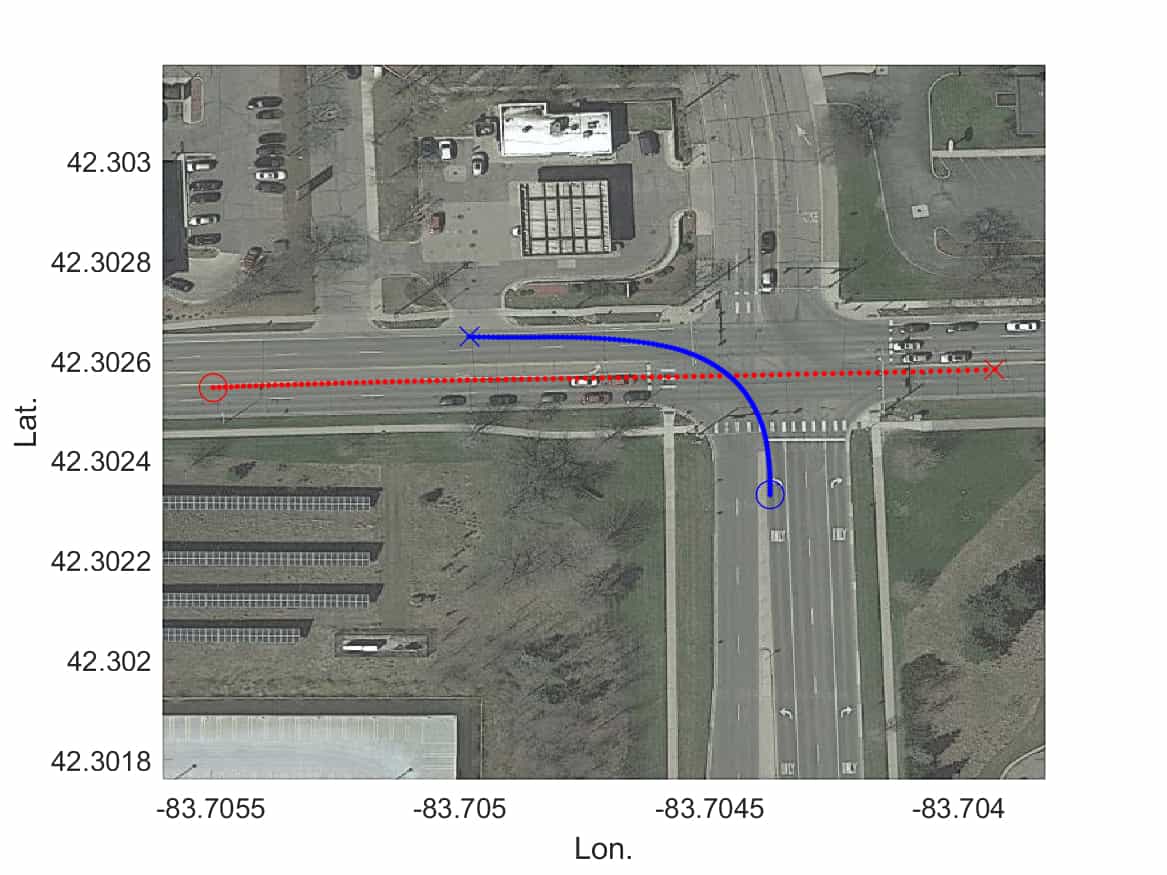}}\hfill
   \subfloat[\label{fig: int5}]{%
       \includegraphics[trim=5.5cm 2.9cm 4.5cm 2cm, clip=true,width=0.49\linewidth]{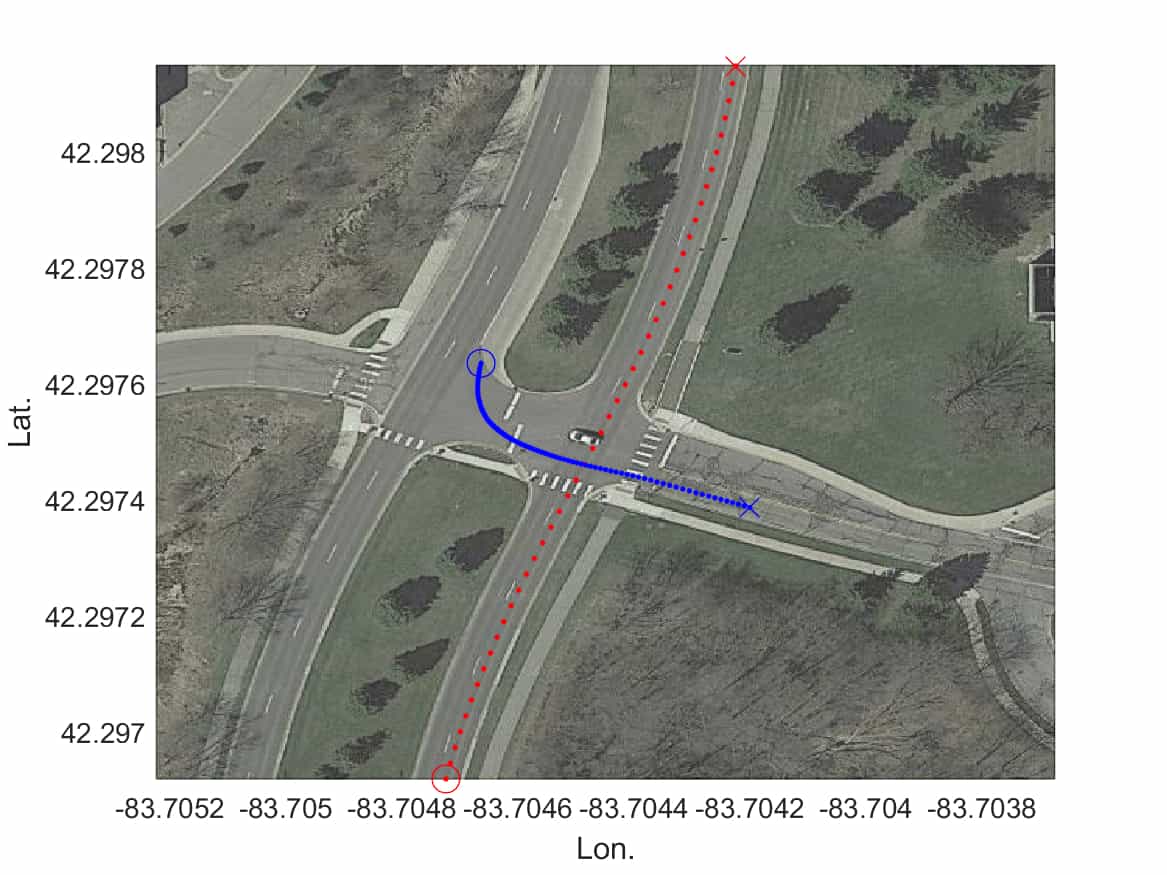}}\hfill
  \subfloat[\label{fig: int6}]{
        \includegraphics[trim=5.5cm 2.9cm 4.5cm 2cm, clip=true,width=0.49\linewidth]{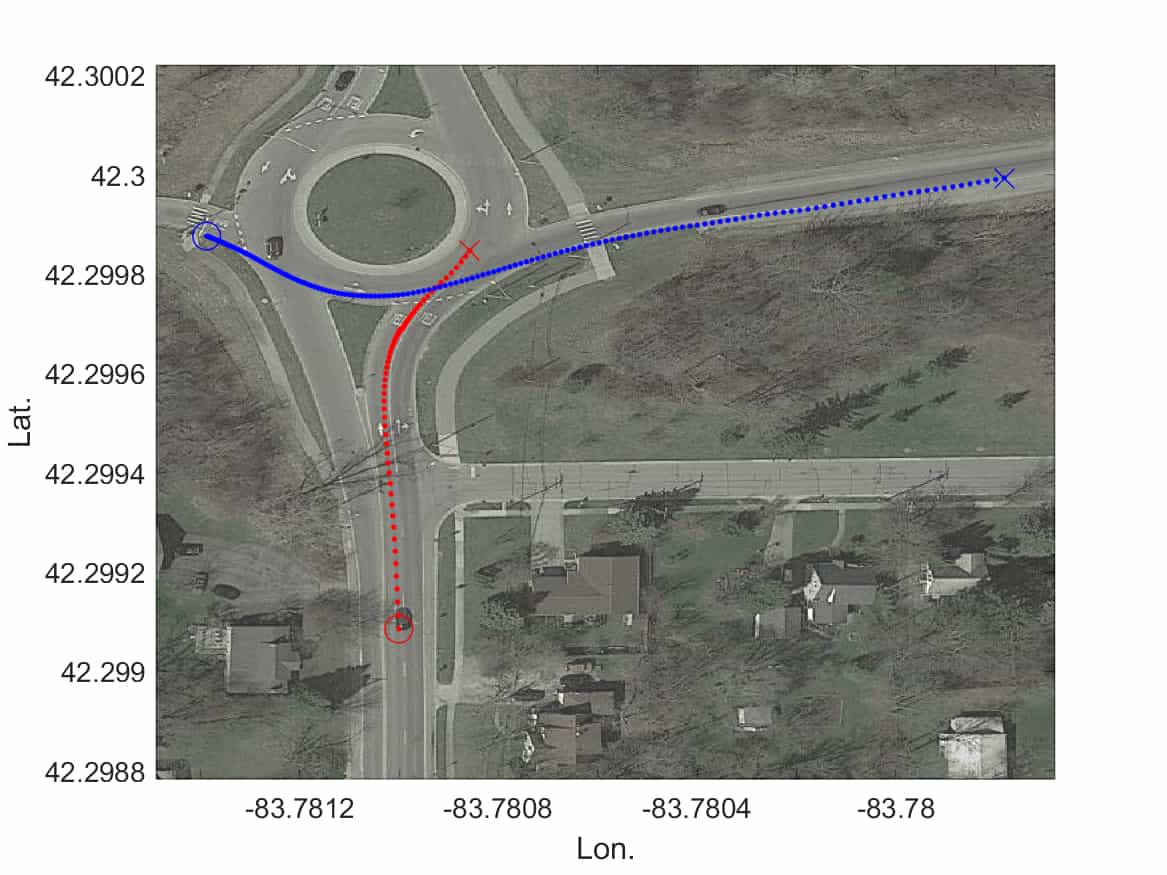}}
  \caption{Cluster of two vehicles encountering at intersections. Circle represents the start positions and cross represents the end position.}
  \label{fig: cluster_intersect} 
\end{figure}

In this section, we will present and compare the cluster results obtained from the two unsupervised learning methods. In real traffic environment, the typical driving encounters mainly consist of four cases: 
\begin{enumerate}
\item Category A: Two vehicles encounter with each other in an intersection.
\item Category B: Two vehicles encounter in the opposite direction of a road.
\item Category C: One vehicle bypasses another vehicle;
\item Category D: Two vehicles interact in a same road (with and without lane changing).
\end{enumerate}
Fig.~\ref{fig: cluster_intersect} shows a sample of cluster results that contains the interaction behavior in an intersection. The developed AE-$k$MC successful clusters the vehicle interactions that fall into the above four categories. 

\begin{figure} [ht]
    \centering
  \subfloat[\label{fig: para1}]{%
       \includegraphics[trim=5.5cm 5.5cm 5cm 3cm, clip=true,width=0.49\linewidth]{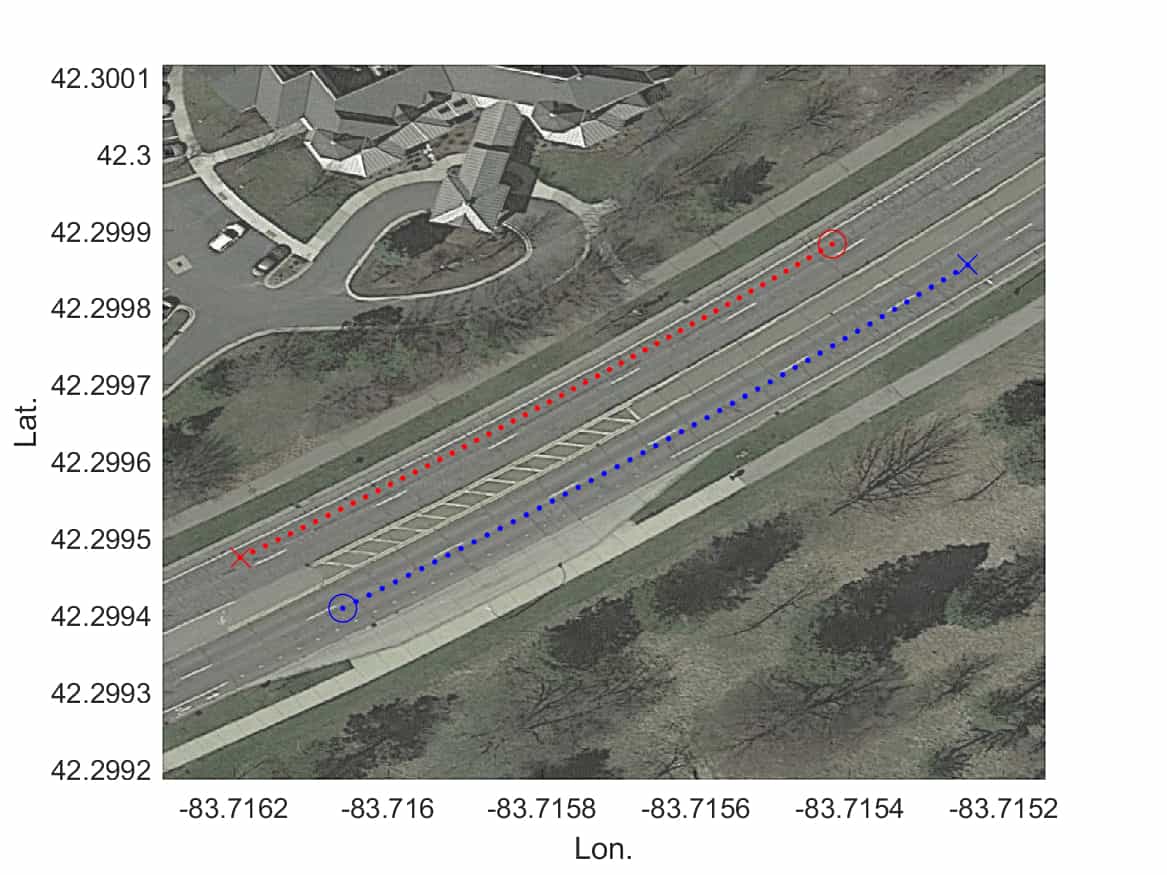}}\hfill
  \subfloat[\label{fig: para2}]{%
        \includegraphics[trim=5.5cm 5.5cm 5cm 3cm, clip=true,width=0.49\linewidth]{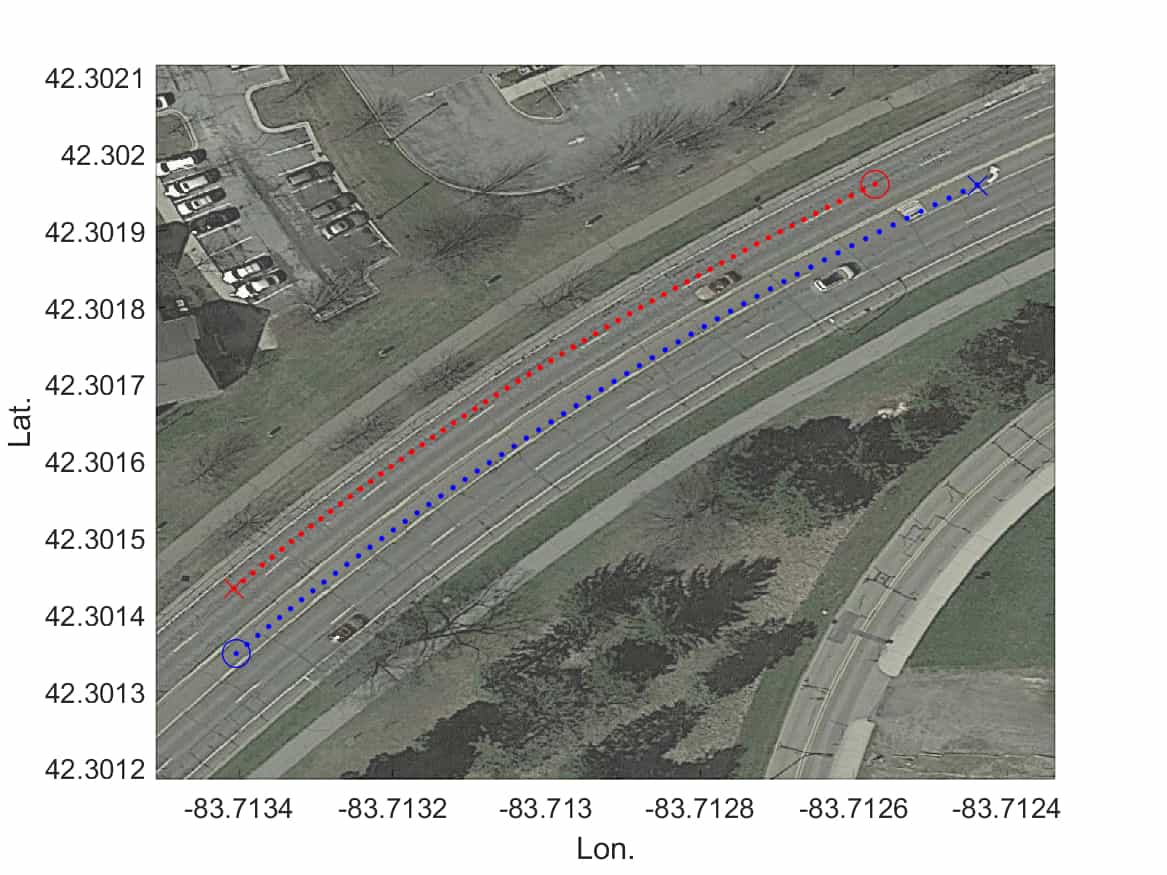}}\hfill
     \subfloat[\label{fig: para3}]{%
       \includegraphics[trim=5.5cm 2.9cm 5cm 2cm, clip=true,width=0.49\linewidth]{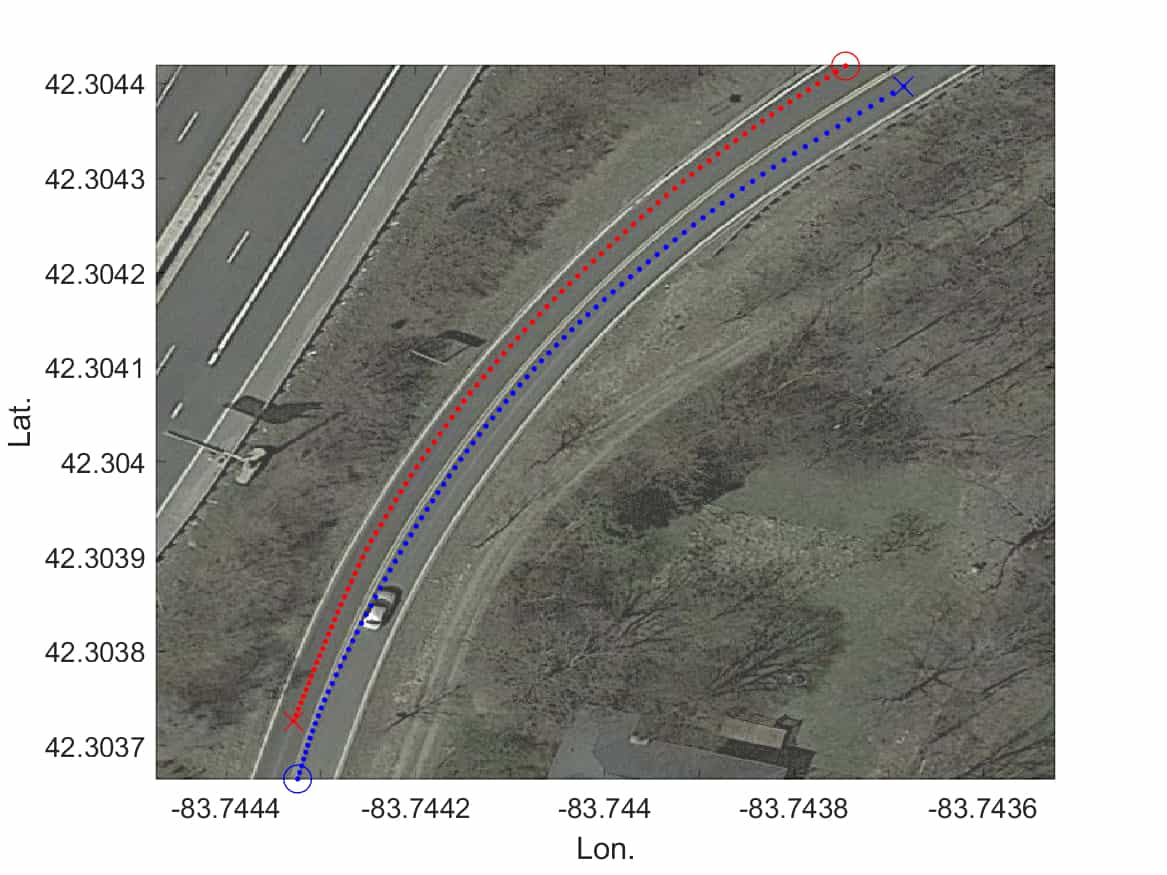}}\hfill
  \subfloat[\label{fig: para4}]{%
        \includegraphics[trim=5.5cm 2.9cm 5cm 2cm, clip=true,width=0.49\linewidth]{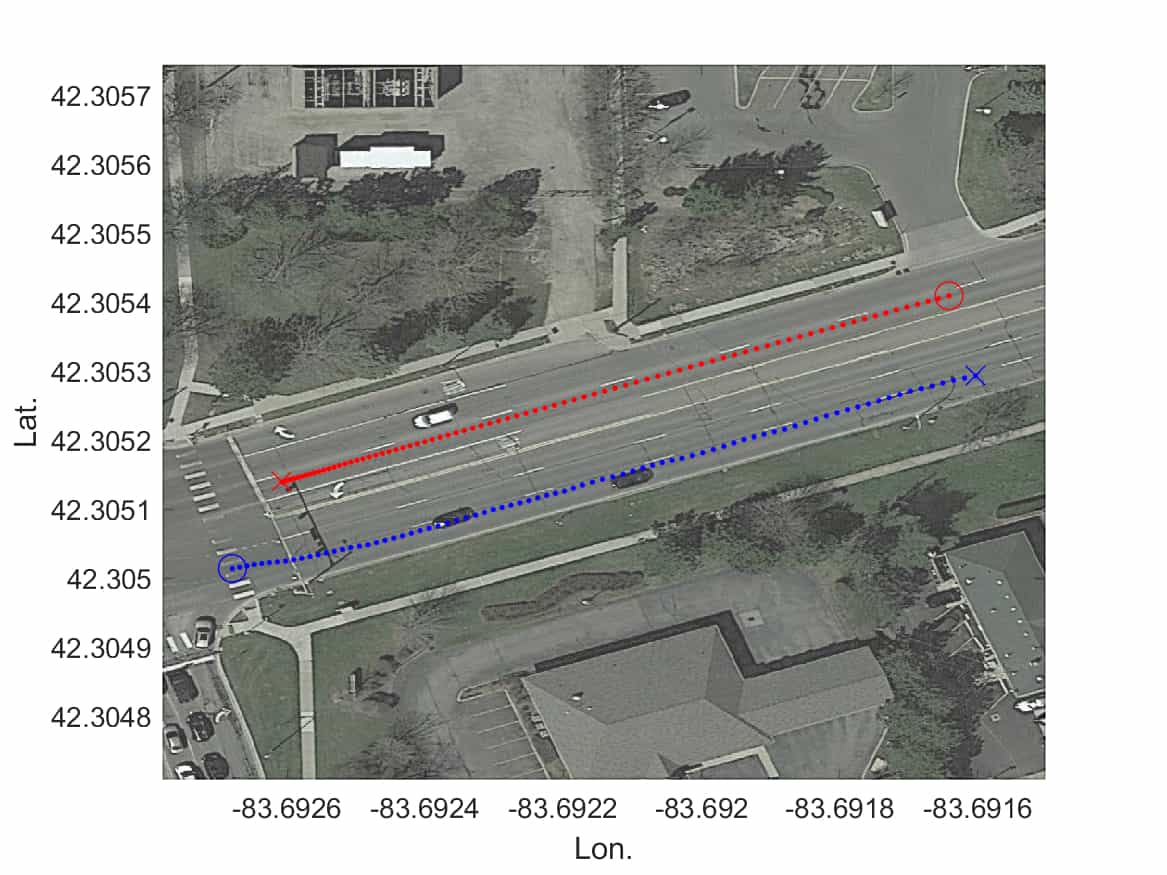}}\hfill
   \subfloat[\label{fig: para5}]{%
       \includegraphics[trim=5.5cm 2.9cm 5cm 2cm, clip=true,width=0.49\linewidth]{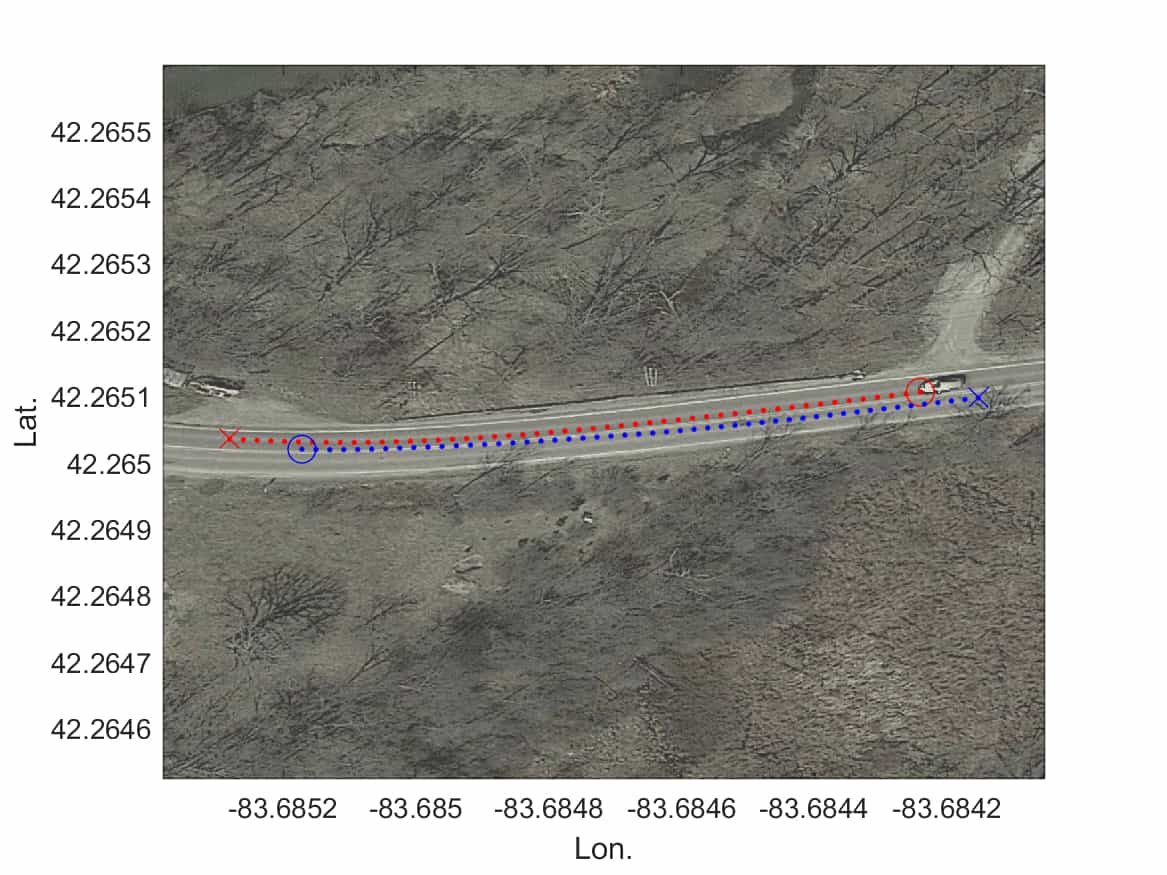}}
    \hfill
  \subfloat[\label{fig: para6}]{%
        \includegraphics[trim=5.5cm 2.9cm 5cm 2cm, clip=true,width=0.49\linewidth]{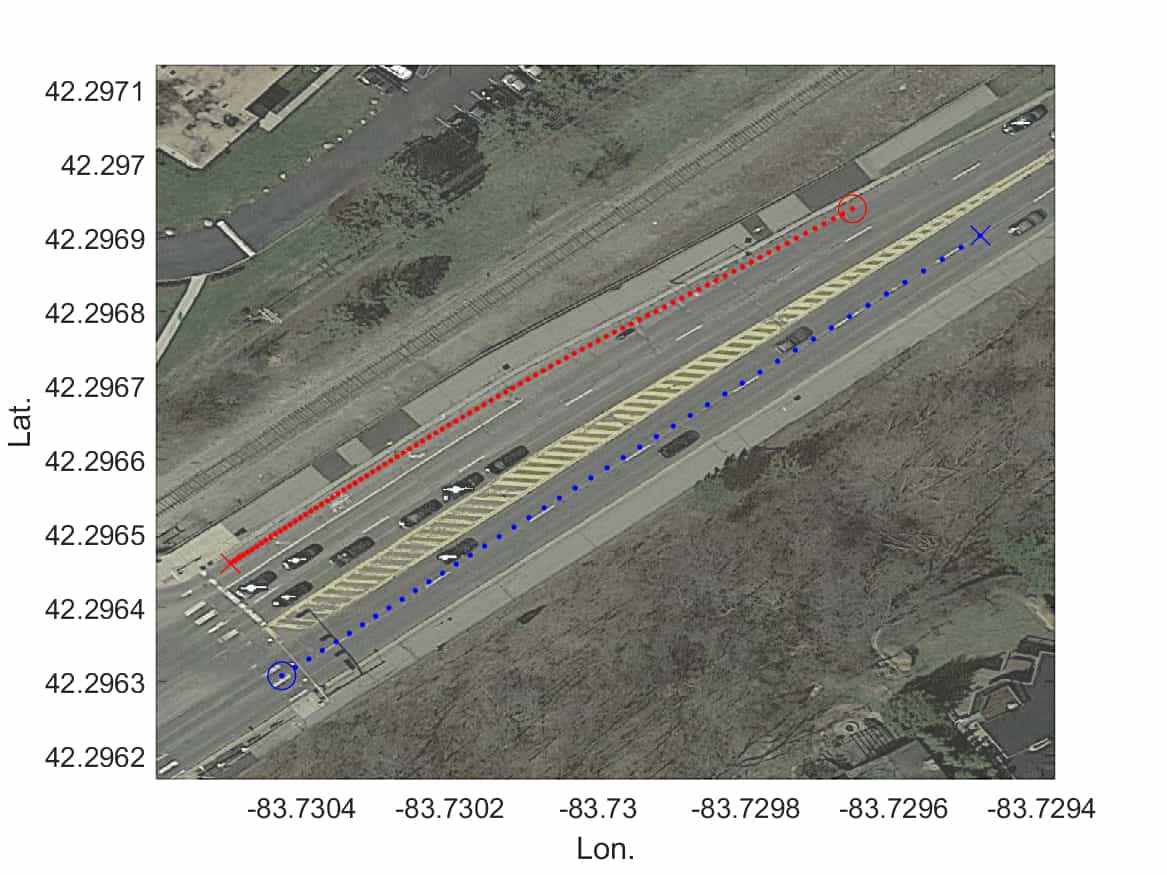}}
  \caption{Clusters of two vehicle encounter on the same road with opposite directions.}
  \label{fig: cluster_parallel} 
\end{figure}

\begin{figure} [ht]
    \centering
  \subfloat[\label{fig: follow1}]{%
       \includegraphics[trim=5.5cm 4.5cm 2.5cm 1cm, clip=true,width=0.49\linewidth]{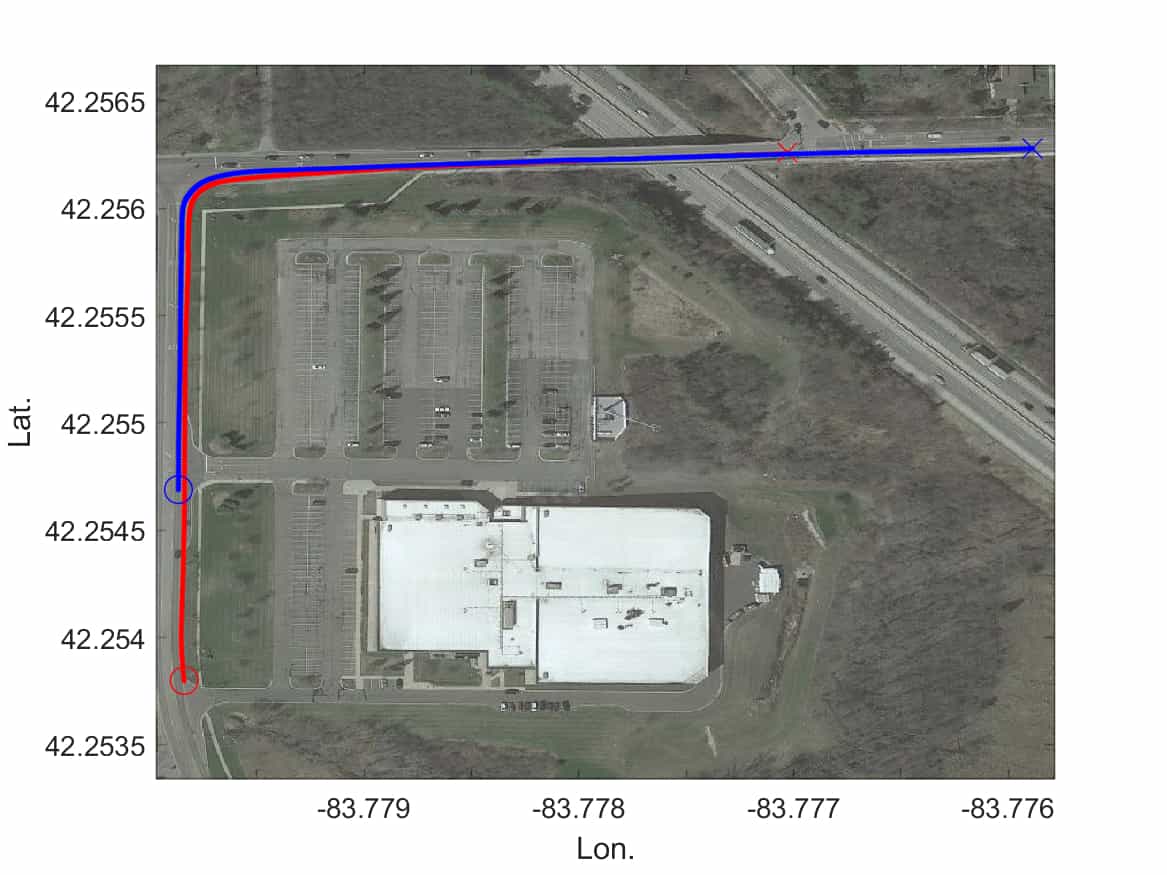}}\hfill
  \subfloat[\label{fig: follow2}]{%
        \includegraphics[trim=5.5cm 4.5cm 2.5cm 1cm, clip=true,width=0.49\linewidth]{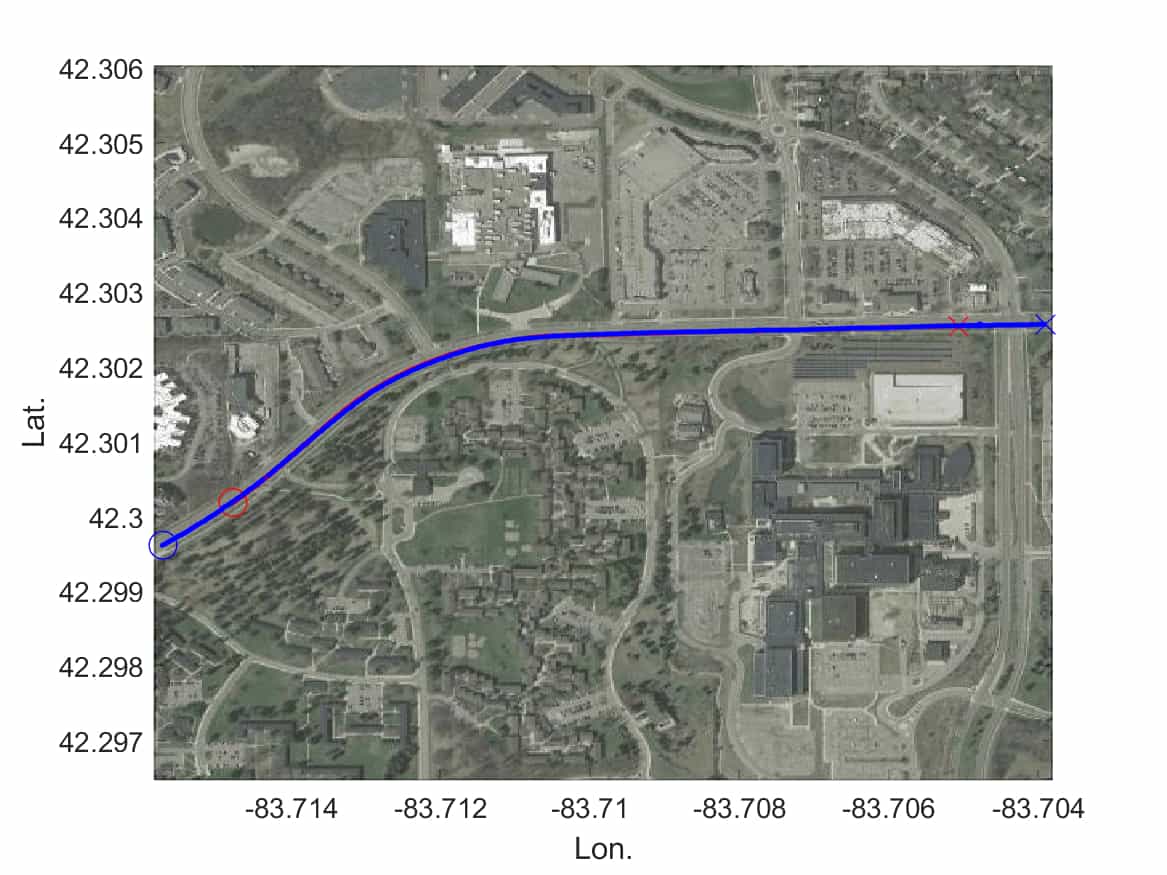}}\hfill
     \subfloat[\label{fig: follow3}]{%
       \includegraphics[trim=5.5cm 4.5cm 3cm 1cm, clip=true,width=0.49\linewidth]{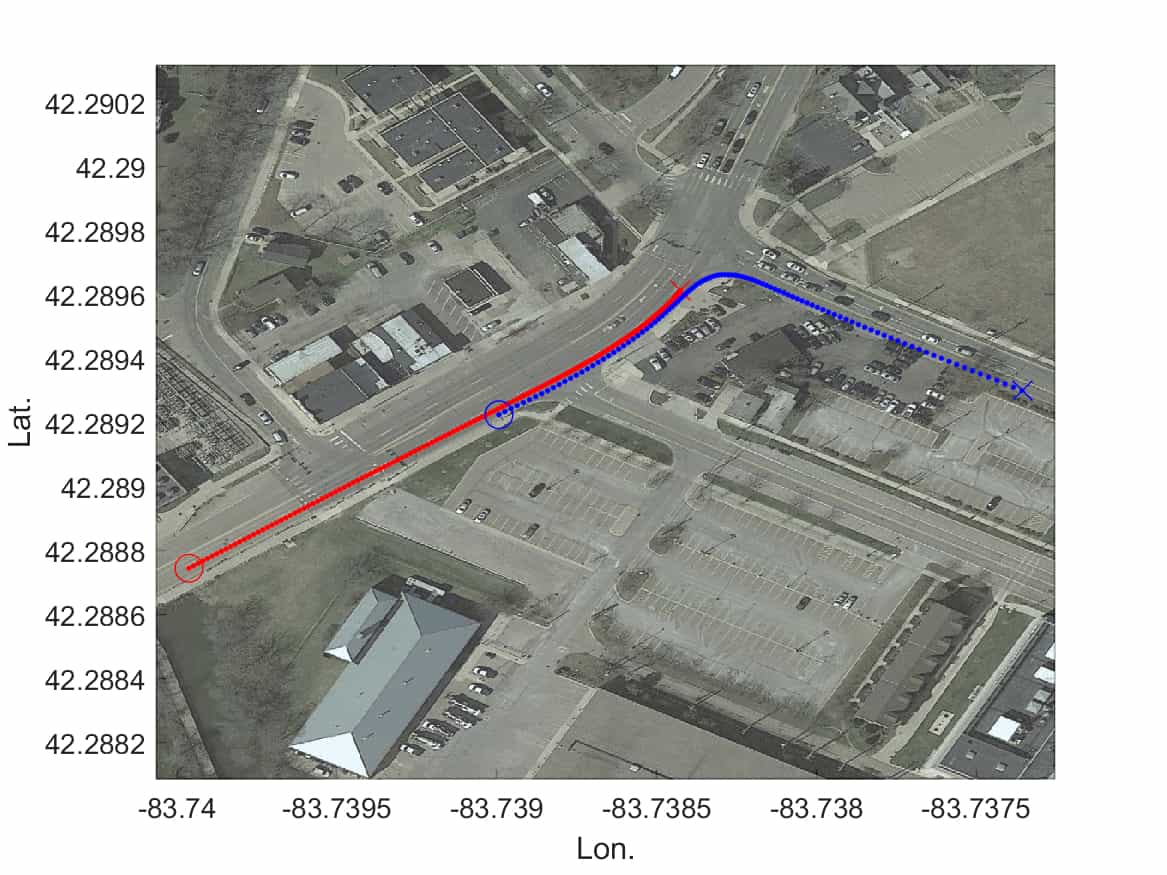}}\hfill
  \subfloat[\label{fig: follow4}]{%
        \includegraphics[trim=5.5cm 4.5cm 3cm 1cm, clip=true,width=0.49\linewidth]{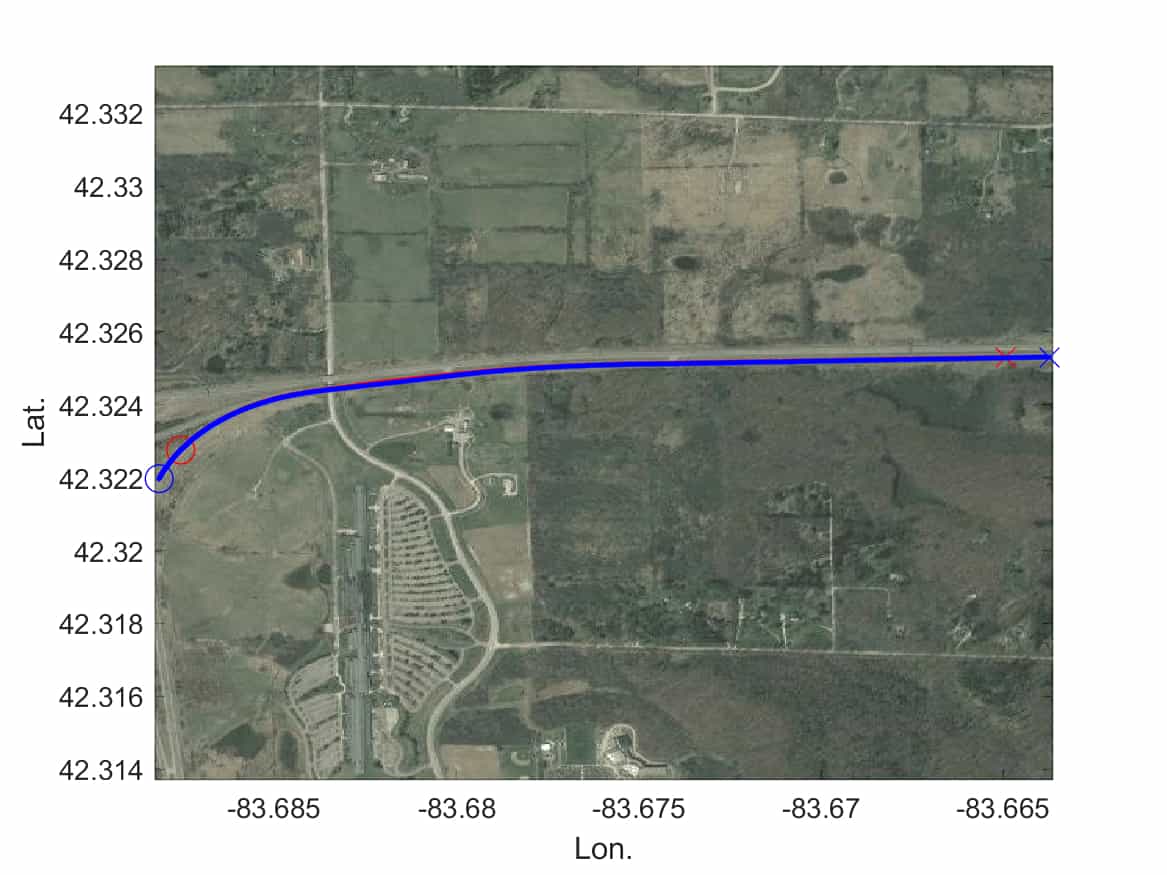}}\hfill
   \subfloat[\label{fig: follow5}]{%
       \includegraphics[trim=5.5cm 3cm 3cm 1cm, clip=true,width=0.49\linewidth]{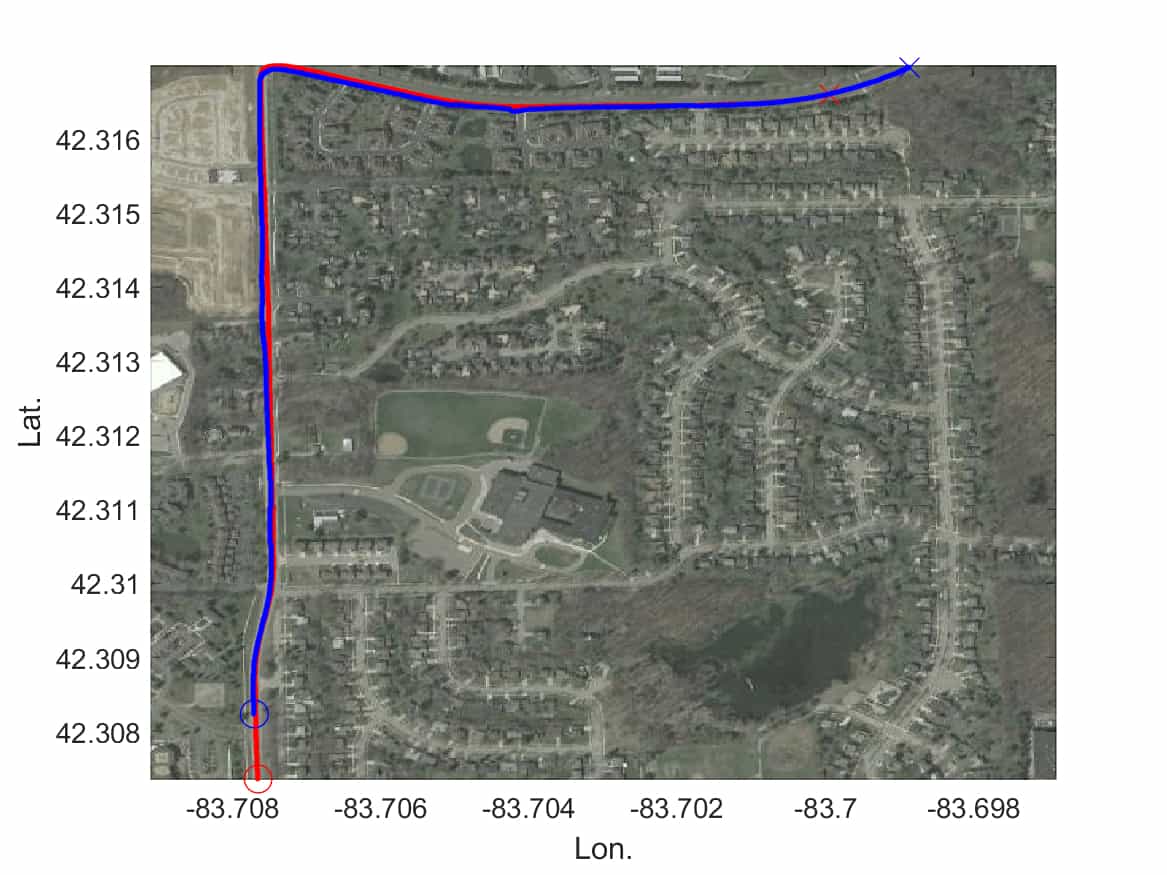}}
    \hfill
  \subfloat[\label{fig: follow6}]{%
        \includegraphics[trim=5.5cm 3cm 3cm 1cm, clip=true,width=0.49\linewidth]{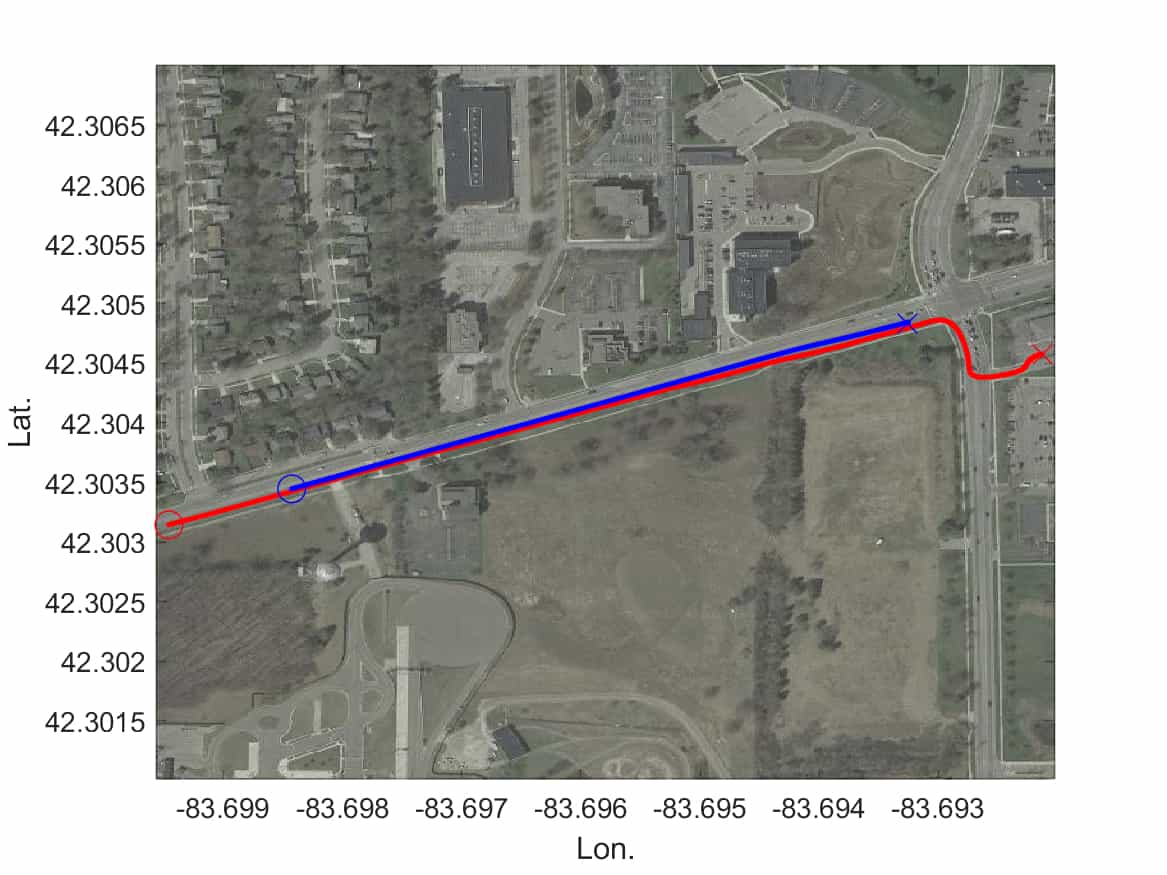}}
  \caption{Cluster of two vehicles encountering on the same road.}
  \label{fig: cluster_follow} 
\end{figure}

The scenario where two vehicles encounter on the same road from opposite directions are shown in Fig.~\ref{fig: cluster_parallel}. Fig.~\ref{fig: cluster_follow} demonstrates two vehicles interact on the same road. Note that the lane changing actions are contained in this cluster but are difficult to be identified from the results. 

\begin{figure} [ht]
    \centering
  \subfloat[\label{fig: bypass1}]{%
       \includegraphics[trim=5.5cm 2.9cm 5cm 2cm, clip=true,width=0.49\linewidth]{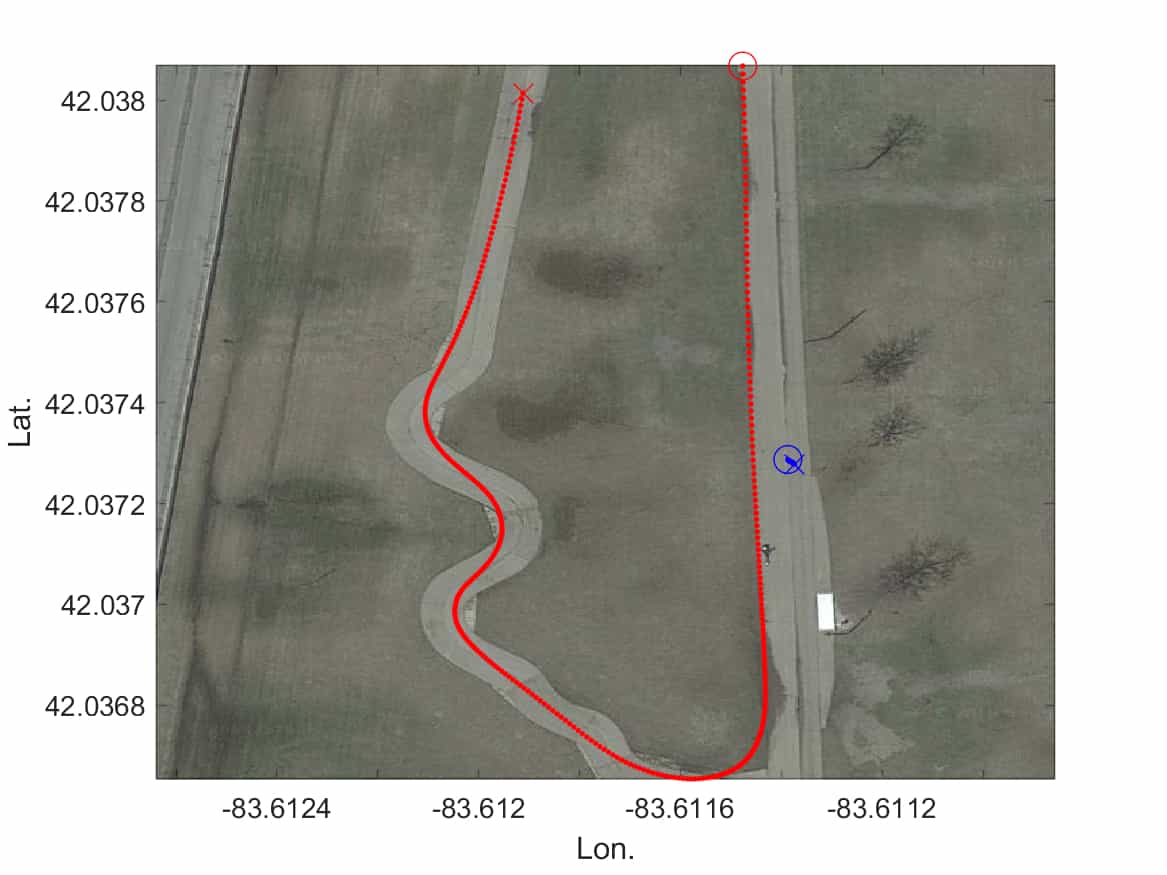}}\hfill
  \subfloat[\label{fig: bypass2}]{%
        \includegraphics[trim=5.5cm 2.9cm 5cm 2cm, clip=true,width=0.49\linewidth]{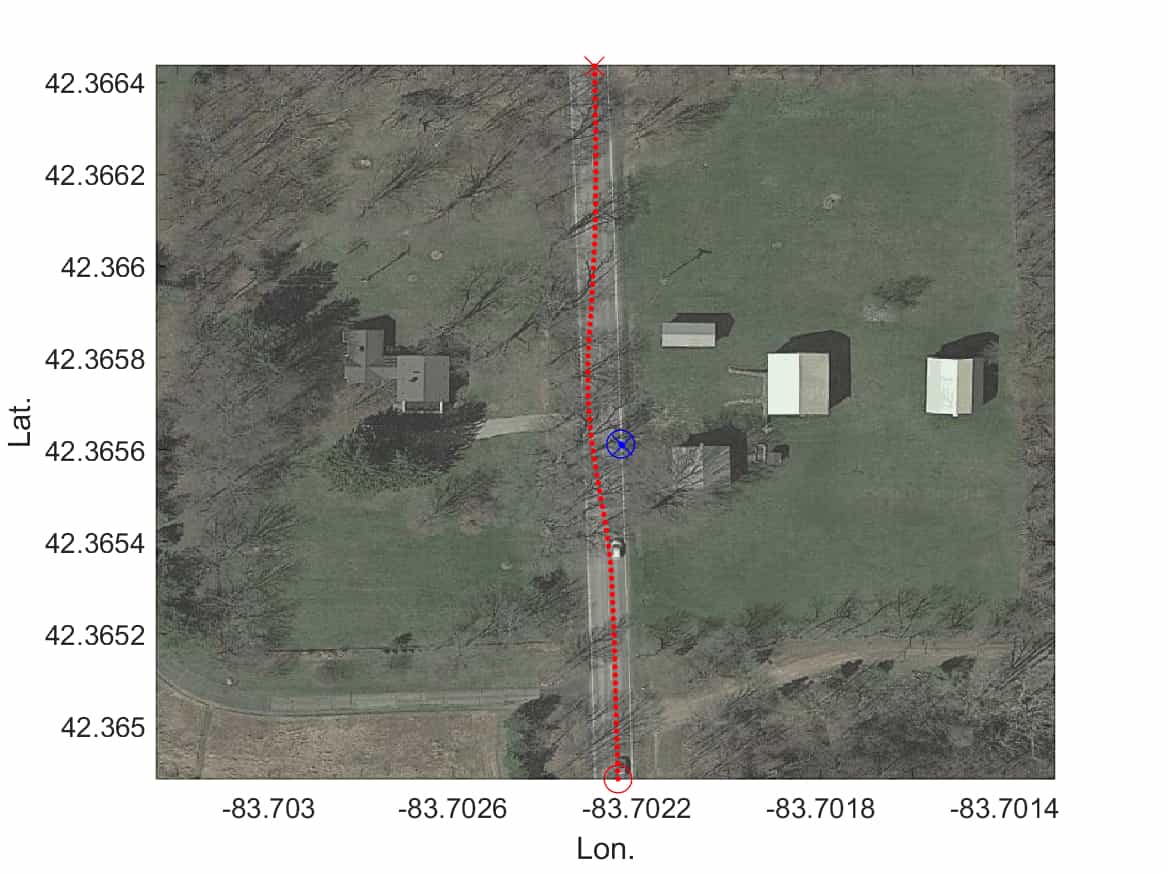}}\hfill
     \subfloat[\label{fig: bypass3}]{%
       \includegraphics[trim=5.5cm 2.9cm 5cm 2cm, clip=true,width=0.49\linewidth]{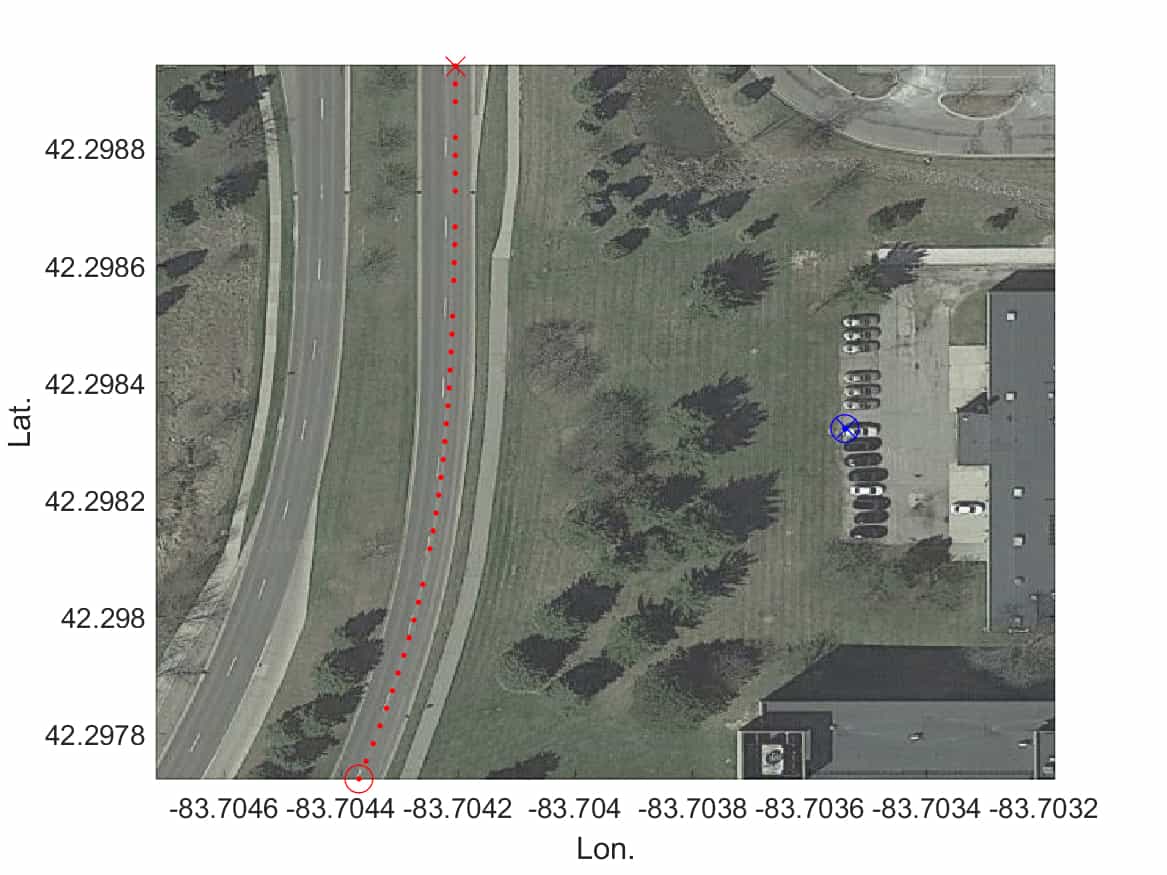}}\hfill
  \subfloat[\label{fig: bypass4}]{%
        \includegraphics[trim=5.5cm 2.7cm 5cm 2cm, clip=true,width=0.49\linewidth]{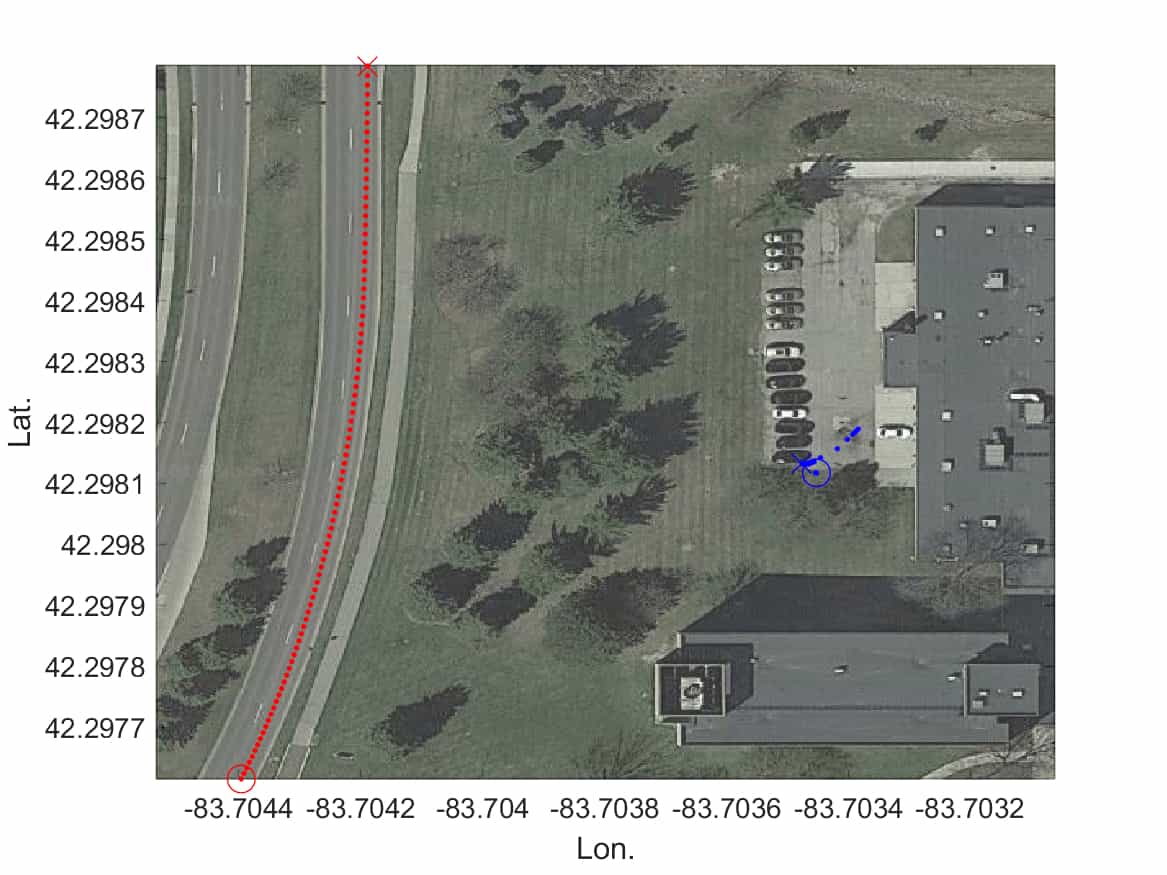}}\hfill
   \subfloat[\label{fig: bypass5}]{%
       \includegraphics[trim=5.5cm 2.9cm 5cm 2cm, clip=true,width=0.49\linewidth]{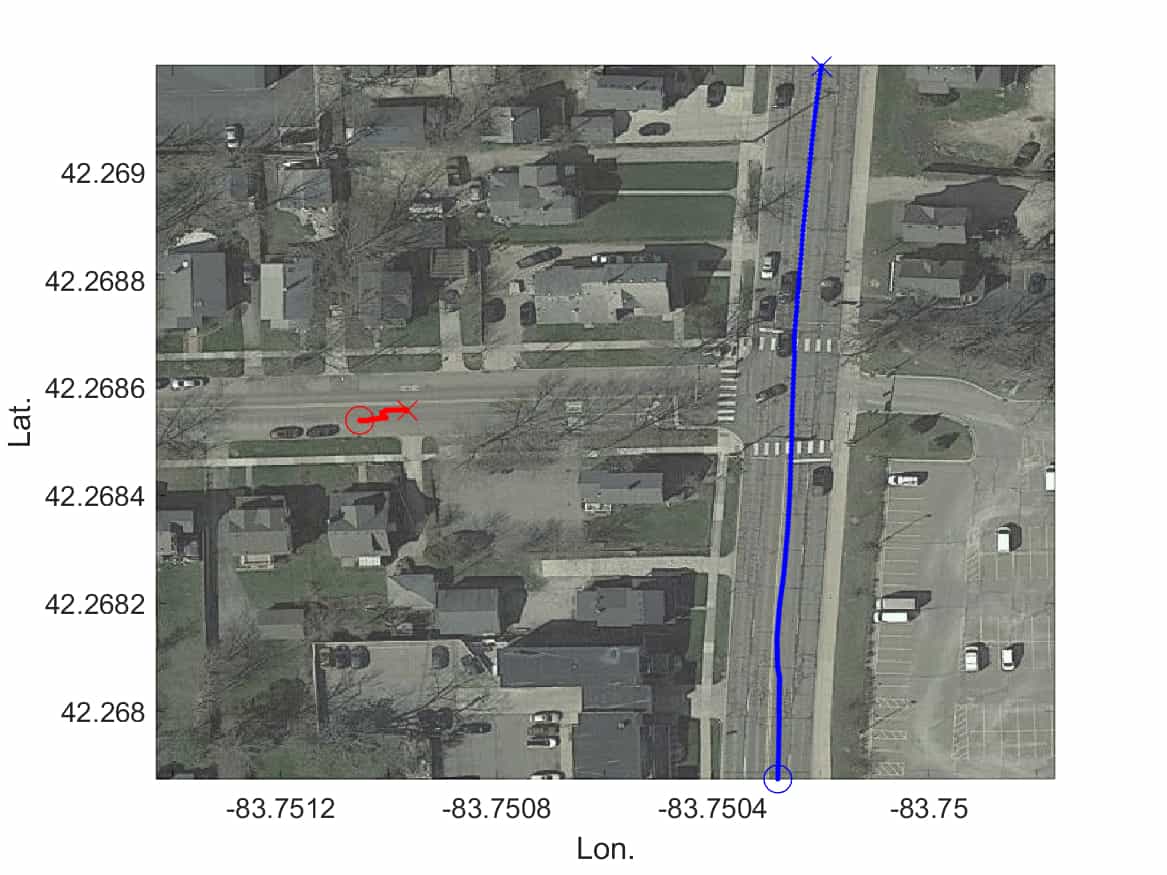}}
    \hfill
  \subfloat[\label{fig: bypass6}]{%
        \includegraphics[trim=5.5cm 2.9cm 5cm 2cm, clip=true,width=0.49\linewidth]{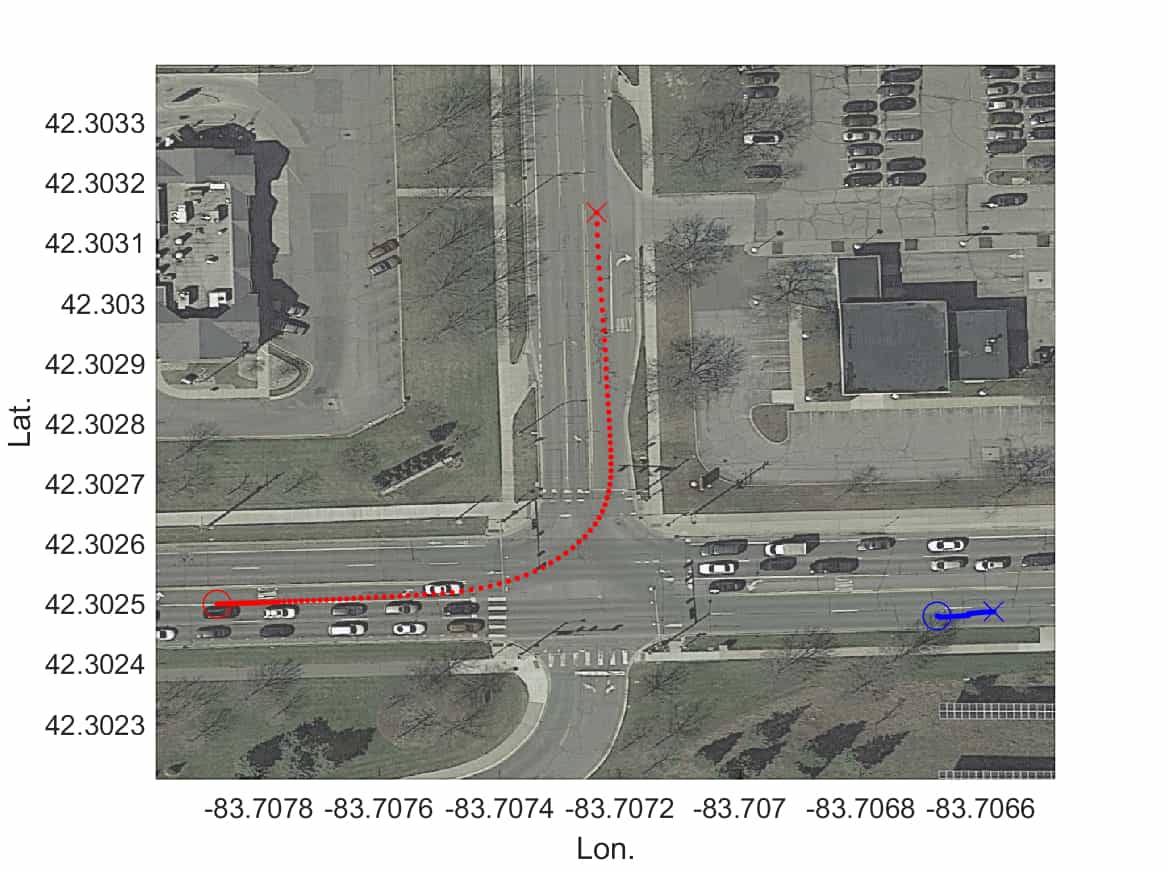}}
  \caption{Cluster of one vehicle passing another vehicle.}
  \label{fig: cluster_bypass} 
\end{figure}

The traffic scenario where one vehicle passes another vehicle is shown in Fig.~\ref{fig: cluster_bypass}. Note that in this case, one vehicle is stationary (Fig.~\ref{fig: bypass1}, Fig.~\ref{fig: bypass2}, Fig.~\ref{fig: bypass3} and Fig.~\ref{fig: bypass4}) or moving but located at far distance from the other vehicle (Fig.~\ref{fig: bypass5} and Fig.~\ref{fig: bypass6}). 

To evaluate the accuracy, we randomly shuffled and re-sampled 100 samples for each cluster obtained from the proposed method, then we manually identified the ones that do not fit into the patterns that most of the samples in the cluster shown. Thus the performance of each cluster is calculated by

\begin{equation}
    \eta = 1 - \frac{n_{abnormal}}{N_{sample}},
\end{equation}
where $n_{abnormal}$ is the number of the abnormal data and $N_{sample} = 100$. The method we proposed can cluster vehicle encounters that have similar geographic features but each cluster still contains abnormal driving encounters. Table~\ref{tab: accuracy} shows the performance of the clusters. The method of only utilizing $k$-means clustering is also evaluated and the results indicate that it can also cluster the above vehicle interaction categories with a lower performance compared with the method uses AE-$k$MC. 

\begin{table}[ht]
\centering
\caption{Clustering performance analysis}
\label{tab: accuracy}
\begin{tabular}{c c c }
\hline
\hline
Cluster   & AE-$k$MC &    $k$-means\\
\hline
 Category A    &  73\%  & 41.6\%   \\
 Category B    &  74.4\%  & 37.9\%\\
 Category C    &  68.2\%   & 47.6\%  \\
 Category D    &  83.8\%   & 78.4\% \\
\hline\hline
\end{tabular}
\end{table}

\section{Conclusion and Future Work}
\label{sec: conclusion}

In this paper, we proposed an unsupervised learning method to cluster vehicle encounter data. More specifically, the auto-encoder was introduced to extract the hidden feature of driving encounters and the extracted features were then grouped using the $k$-means clustering method. The proposed AE-$k$MC method finally obtains five main typical types of vehicle encounters, including 1) two vehicles intersect with each other;  two vehicles merge; 3) two vehicles encounter in the opposite direction of a road; 4) one vehicle bypasses another vehicle; 5) two vehicles interact in a same road (with and without lane changing). We also compared the developed AE-$k$MC method with the $k$-means clustering method and the result shows that our developed AE-$k$MC method outperforms the $k$-means clustering method. 

In future work, we will focus on improving the accuracy of the unsupervised clustering approaches, including evaluating different types of the auto-encoders and clustering approaches. Moreover, target vehicle encounter extraction, for example, vehicle interactions in roundabout traffic environment, will be investigated.

\section*{Acknowledgment}
\label{sec: ack}

Toyota Research Institute (``TRI") provided funds to assist the authors with their research but this article solely reflects the opinions and conclusions of its authors and not TRI or any other Toyota entity.  

\ifCLASSOPTIONcaptionsoff
  \newpage
\fi

\bibliographystyle{IEEEtran}
\bibliography{IEEEabrv,References}

\end{document}